\documentclass[twoside,11pt]{article}

\usepackage{times}
\usepackage{fullpage}
\usepackage{hyperref}       
\hypersetup{
    colorlinks,
    linkcolor=black, 
    citecolor=black, 
    linktoc=all 
  }

\RequirePackage{amsmath, amssymb}%
\usepackage{mathtools}%
\usepackage{dsfont}
\usepackage{booktabs,multirow,array}
\usepackage{mathrsfs}
\usepackage{subfigure} 
\usepackage{listings}
\usepackage{scrextend}
\usepackage[linesnumbered,ruled,vlined]{algorithm2e}

\DeclareMathOperator{\argmin}{argmin}
\DeclareMathOperator{\prox}{{prox}}
\DeclareMathOperator{\sgn}{sign}
\DeclareMathOperator{\diag}{diag}
\DeclareMathOperator{\TV}{TV}
\DeclareMathOperator{\bina}{bina}

\newtheorem{assumption}{Assumption}{\bf}{\rm}
\newtheorem{proposition}{Proposition}{\bf}{\rm}
\newtheorem{theorem}{Theorem}{\bf}{\rm}
\newtheorem{lemma}{Lemma}{\bf}{\rm}

\newcommand{\inr}[1]{\langle #1 \rangle}%
\newcommand{\norm}[1]{\|#1\|}
\newcommand{\R}{{\mathbb{R}}}
\newcommand{\N}{{\mathbb{N}}}
\newcommand{\E}{\mathds{E}} 

\renewcommand{\P}{\mathds{P}}
\newcommand{\bigO}{O}
\newcommand{\cJ}{\mathcal{J}}
\newcommand{\XB}{\boldsymbol{X}^{{B}}}%
\newcommand{\bX}{{\boldsymbol X}}%
\newcommand{\by}{{\boldsymbol{y}}}%
\newcommand{\cY}{{\mathcal{Y}}}%
\newcommand{\Pyx}{\P_{\mathbf y|X}}

\newcommand{\ind}[1]{\mathbf 1_{#1}}
\newcommand{\bul}{\bullet}%

\begin{document}

\title{Binarsity: a penalization for one-hot encoded features in linear supervised learning}

\author{
  Mokhtar Z. Alaya\footnote{LPSM, CNRS UMR 8001, Sorbonne University, Paris France}
  \and Simon Bussy\footnotemark[1]
  \and St\'ephane Ga\"iffas\footnote{LPSM, CNRS UMR 8001, Universit\'e Paris Diderot, Paris, France}
  \and Agathe Guilloux\footnote{LaMME, UEVE and UMR 8071, Universit\'e Paris Saclay, Evry, France}
}

\maketitle

\begin{abstract}
This paper deals with the problem of large-scale linear supervised learning in settings where a large number of continuous features are available.
We propose to combine the well-known trick of one-hot encoding of continuous features with a new penalization called \emph{binarsity}.
In each group of binary features coming from the one-hot encoding of a single raw continuous feature, this penalization uses total-variation regularization together with an extra linear constraint.
This induces two interesting properties on the model weights of the one-hot encoded features: they are piecewise constant, and are eventually block sparse.
Non-asymptotic oracle inequalities for generalized linear models are proposed.
Moreover, under a sparse additive model assumption, we prove that our procedure matches the state-of-the-art in this setting.
Numerical experiments illustrate the good performances of our approach on several datasets.
It is also noteworthy that our method has a numerical complexity comparable to standard $\ell_1$ penalization.

\end{abstract}

\section{Introduction}

In many applications, datasets used for linear supervised learning contain a large number of continuous features, with a large number of samples.
An example is web-marketing, where features are obtained from bag-of-words scaled using tf-idf~\cite{russell2013mining}, recorded during the visit of users on websites.
A well-known trick~\cite{WuCog2012,LiuHussTanDas2002} in this setting is to replace each raw continuous feature by a set of binary features that one-hot encodes the interval containing it, among a list of intervals partitioning the raw feature range.
This improves the linear decision function with respect to the raw continuous features space, and can therefore improve prediction.
However, this trick is prone to over-fitting, since it increases significantly the number of features.

\paragraph{A new penalization.}

To overcome this problem, we introduce a new penalization called \emph{binarsity}, that penalizes the model weights learned from such grouped one-hot encodings (one group for each raw continuous feature).
Since the binary features within these groups are naturally ordered, the binarsity penalization combines a group total-variation penalization, with an extra linear constraint in each group to avoid collinearity between the one-hot encodings.
This penalization forces the weights of the model to be as constant (with respect to the order induced by the original feature) as possible within a group, by selecting a minimal number of relevant cut-points.
Moreover, if the model weights are all equal within a group, then the full block of weights is zero, because of the extra linear constraint.
This allows to perform raw feature selection.

\paragraph{High-dimensional linear supervised learning.}

To address the high-dimensionality of features, sparse linear inference is now an ubiquitous technique for dimension reduction and variable selection, see for instance~\cite{BuhVan-11} and \cite{ESL} among many others. 
The principle is to induce sparsity (large number of zeros) in the model weights, assuming that only a few features are actually helpful for the label prediction.
The most popular way to induce sparsity in model weights is to add 
a $\ell_1$-penalization (Lasso) term to the 
goodness-of-fit~\cite{Tib-96}.
This typically leads to sparse parametrization of models, with a level of sparsity that depends on the strength of the penalization.
Statistical properties of $\ell_1$-penalization have been extensively investigated, see for instance~\cite{KniFu-00,zhaoconsistency2006, bunea2007oracle,BicRitTsy-09} for linear and generalized linear models and~\cite{donoho12001,Donoho02optimallysparse,candes2008a,candes2008b} for compressed sensing, among others. 

However, the Lasso ignores ordering of features.
In~\cite{TibRosZhuKni-05}, a structured sparse penalization is proposed, known as fused Lasso, which provides superior performance in recovering the true model in such applications where features are ordered in some meaningful way.
It introduces a mixed penalization using a linear combination of the $\ell_1$-norm and the total-variation penalization, thus enforcing sparsity in both the weights and their successive differences. 
Fused Lasso has achieved great success in some applications such as comparative genomic hybridization~\cite{rapaport2008}, image denoising~\cite{FriHasHofTib-07}, and prostate cancer analysis~\cite{TibRosZhuKni-05}.

\paragraph{Features discretization and cuts.}

For supervised learning, it is often useful to encode the input features in a new space to let the model focus on the relevant areas~\cite{WuCog2012}.
One of the basic encoding technique is \emph{feature discretization} or \emph{feature quantization}~\cite{LiuHussTanDas2002} that partitions the range of a continuous feature into intervals and relates these intervals with meaningful labels. 
Recent overviews of discretization techniques can be found in~\cite{LiuHussTanDas2002} or~\cite{GarLueSaeLopHer2013}.

Obtaining the optimal discretization is a NP-hard problem~\cite{ChlNgu1998}, and an approximation can be easily obtained using a greedy approach, as proposed in decision trees: CART~\cite{BreFriOlsSto-84} and C4.5~\cite{Qui-93}, among others, that sequentially select pairs of features and cuts that minimize some purity measure (intra-variance, Gini index, information gain are the main examples).
These approaches build decision functions that are therefore very simple, by looking only at a single feature at a time, and a single cut at a time.
Ensemble methods (boosting~\cite{lugosi2004bayes}, random forests~\cite{breiman2001random}) improve this by combining such decisions trees, at the expense of models that are harder to interpret.

\paragraph{Main contribution.}

This paper considers the setting of linear supervised learning. The main contribution of this paper is the idea  to use a total-variation penalization, with an extra linear constraint, on the weights of a generalized linear  model trained on a binarization of the raw continuous features, leading to a procedure that selects multiple cut-points per feature, looking at all features simultaneously.
Our approach therefore increases the capacity of the considered generalized linear model: several weights are used for the binarized features instead of a single one for the raw feature. 
This leads to a more flexible decision function compared to the linear one: when looking at the decision function as a function of a single raw feature, it is now piecewise constant instead of linear, as illustrated in Figure~\ref{figure-DecisionFunc} below.

\paragraph{Organization of the paper.}

The proposed methodology is described in Section~\ref{section:methodology}.
Section~\ref{sec:theoretical_results} establishes an oracle inequality for generalized linear models and provides a convergence rate for our procedure in the particular case of a sparse additive model.
Section~\ref{section:btv-experiments} highlights the results of the method on various datasets and compares its performances to well known classification algorithms.
Finally, we discuss the obtained results in 
Section~\ref{section:discussion}.

\paragraph*{Notations.}

Throughout the paper, for every $q > 0,$ we denote by $\norm{v}_q$ the usual $\ell_q$-quasi norm of a vector $v \in \R^m,$ namely
 $\norm{v}_q =(\sum_{k=1}^m|v_k|^q)^{1/q}$,  and $\norm{v}_\infty = \max_{k=1, \ldots, m}|v_k|$. We also denote $\norm{v}_0 = |\{k : v_k \neq 0\}|$, where $|A|$ stands for the cardinality of a finite set $A$.
 For $u, v \in \R^m$, we denote by $u \odot v$ the Hadamard product $u\odot v =(u_1v_1, \ldots, u_mv_m)^\top.$ 
For any $u \in \R^m$ and any $L \subset \{1, \ldots, m\},$ we denote $u_L$ as the vector in $\R^m$ satisfying $(u_L)_k = u_k$ for $k \in L$ and $(u_L)_k = 0$ for $ k \in L^\complement =  \{1, \ldots, m\}\backslash L$.
We write, for short, $\mathbf{1}$  (resp. $\mathbf{0}$) for the vector of $\R^m$ having all coordinates equal to one (resp. zero). 
Finally, we denote by $\sgn(x)$ the set of sub-differentials of the function $x \mapsto |x|$, namely $\sgn(x) = \{ 1\}$ if $x > 0$, $\sgn(x) = \{ -1 \}$ if $x < 0$ and $\sgn(0) = [-1, 1]$.

\section{The proposed method}
\label{section:methodology}

Consider a supervised training dataset $(x_i, y_i)_{i=1, \ldots, n}$ containing features 
$x_i = [x_{i,1} \cdots x_{i,p}]^\top \in \R^p$ and labels
 $y_i \in \cY \subset \R$, that are independent and identically distributed samples of $(X, Y)$ with unknown distribution $\P$.
Let us denote $\bX = [x_{i,j}]_{1 \leq i \leq n; 1 \leq j \leq p}$ the $n \times p$ features matrix vertically stacking the $n$ samples of $p$ raw features. Let $\bX_{\bullet, j}$ be the $j$-th feature column of~$\bX$.

\paragraph{Binarization.}

The binarized matrix $\XB$ is a matrix with an extended number $d > p$ of columns, where the $j$-th column $\bX_{\bullet, j}$ is replaced by $d_j \geq 2$ columns $\XB_{\bullet, j, 1}, \ldots, \XB_{\bullet, j, d_j}$ containing only zeros and ones. Its $i$-th row is written 
\begin{equation*}
x_i^B = [x^B_{i,1,1} \cdots x^B_{i,1,d_1} x^B_{i,2,1} \cdots x^B_{i,2,d_2} \cdots x^B_{i,p,1} 
\cdots x^B_{i,p, d_p}]^\top \in \R^d,
\end{equation*}
where $d = \sum_{j=1}^p d_j$.
In order to simplify the presentation of our results, we assume in the paper that all raw 
features $\bX_{\bullet, j}$ are continuous, so that they are transformed using the following one-hot encoding.
For each raw feature $j$, we consider a partition of intervals $I_{j,1}, \ldots, I_{j, d_j}$ of 
$\text{range}(\bX_{\bullet, j})$, namely satisfying $\cup_{k=1}^{d_j}I_{j,k} = \text{range}(\bX_{\bullet, j})$ and $I_{j,k} \cap I_{j,k'} = \varnothing$ for $k \neq k'$ and define
\begin{equation*}
  x^B_{i, j, k} =
  \begin{cases}
    1 &\text{ if } x_{i,j} \in I_{j, k}, \\
    0 & \text{ otherwise}
  \end{cases}
\end{equation*}
for $i=1, \ldots, n$, $j=1, \ldots, p$ and $k=1, \ldots, d_j$. 
An example is interquantiles intervals, namely
$I_{j, 1} = \big[ q_j(0), q_j(\frac {1}{d_j})\big]$ and
$I_{j, k} = \big(q_j(\frac {k-1}{d_j}) , q_j(\frac {k}{d_j}) \big]$ for $k=2, \ldots, d_j$, where $q_j(\alpha)$ denotes a quantile of order $\alpha \in [0, 1]$ for $\bX_{\bullet, j}$.
In practice, if there are ties in the estimated quantiles for a given feature, we simply choose the set of ordered unique values to construct the intervals.
This principle of binarization is a well-known trick~\cite{GarLueSaeLopHer2013}, that allows to improve over the linear decision function with respect to the raw feature space: it uses a larger number of model weights, for each
interval of values for the feature considered in the binarization.
If training data contains also unordered qualitative features, one-hot encoding with $\ell_1$-penalization can be used for instance.

\paragraph{Goodness-of-fit.}

Given a loss function $\ell : \cY \times \R \rightarrow \R$, we consider the goodness-of-fit term 
\begin{equation}
  \label{eq:gof}
  R_n(\theta) =  \frac 1n \sum_{i=1}^n \ell(y_i, m_\theta(x_i)),
\end{equation}
where $m_\theta(x_i) = \theta^\top x_i^B$ and $\theta \in \R^d$ where we recall that
 $d = \sum_{j=1}^p d_j$. We then have $\theta = [\theta_{1, \bullet}^\top \cdots 
 \theta_{p,\bullet}^\top]^\top$, with $\theta_{j,\bullet}$ corresponding to the group of coefficients weighting the binarized raw $j$-th feature.
We focus on generalized linear models~\cite{green1994}, where the conditional distribution $Y | X = x$ is assumed to be from a one-parameter exponential family distribution with a density of the form
\begin{equation}
\label{distribut-glm}
  y | x \mapsto f^0(y | x) = \exp\Big(\frac{ym^0(x) - b(m^0(x))}{\phi} + c(y,\phi)\Big),
\end{equation}
with respect to a reference measure which is either the Lebesgue measure (e.g. in the Gaussian case) or the counting measure (e.g. in the logistic or Poisson cases), leading to a loss function of the form
\begin{equation*}
  \ell\big(y_1, y_2) = - y_1 y_2 + b(y_2).
\end{equation*}
The density described in~\eqref{distribut-glm} encompasses several distributions, see Table~\ref{table:glm}.
The functions $b(\cdot)$ and $c(\cdot)$ are known, while the natural parameter function $m^0(\cdot)$ is unknown.
The dispersion parameter $\phi$ is assumed to be known in what follows.
It is also assumed that $b(\cdot)$ is  three times continuously differentiable.
It is standard to notice that
\begin{equation*}
  \E[Y|X=x] = \int yf^0(y | x) dy = b'(m^0(x)),
\end{equation*}
where $b'$ stands for the derivative of $b$. 
This formula explains how $b'$ links the conditional expectation to the unknown $m^0$.
The results given in Section~\ref{sec:theoretical_results} rely on the following Assumption.

\begin{assumption}
  \label{ass:glm}
  Assume that $b$ is three times continuously differentiable, that there is $C_b > 0$ such that $|b'''(z)| \leq C_b |b''(z)|$ for any $z \in \R$ and that there 
  exist constants $C_n > 0$ and $0 < L_n \leq U_n$ such that $C_n = \max_{i=1, \ldots, n}|m^0(x_i)| < \infty$ and $L_n \leq \max_{i=1, \ldots, n} b''\big(m^0(x_i)\big) \leq  U_n.$ 
\end{assumption}
This assumption is satisfied for most standard generalized linear models. 
In Table~\ref{table:glm}, we list some standard examples that fit in this framework, see also~\cite{vandegeer2008} and~\cite{rigollet2012}.

\begin{table}[htb]
\centering
\begin{tabular}{ccccccccc}
\toprule
 Model & $\phi$ & $b(z)$ & $b'(z)$ & $b''(z)$ & $b'''(z)$ & $C_b$ & $L_n$ & $U_n$\\
 \midrule
 Normal & $\sigma^2$ & $\frac{z^2}{2}$ & $z$ & $1$ & $0$ & $0$ & $1$ & $1$ \\
 Logistic & $1$ & $\log(1 + e^z)$&$\frac{e^z}{1+e^z}$ & $\frac{e^z}{(1+e^z)^2}$ &$\frac{1 - e^z}{1+e^z} b''(z)$ & 2 & $\frac{e^{C_n}}{(1 + e^{C_n})^2}$ & $\frac{1}{4}$\\
 Poisson & $1$ &  $e^z$ & $e^z$ & $e^z$ & $b''(z)$ & 1 & $e^{-C_n}$  & $e^{C_n}$\\
 \bottomrule
\end{tabular}
\caption[table]{\small Examples of standard distributions that fit in the considered setting of generalized linear models, with the corresponding constants in Assumption~\ref{ass:glm}.}
\label{table:glm}
\end{table}

\paragraph{Binarsity.}

Several problems occur when using the binarization trick described above:
\begin{enumerate}
  \item[(P1)] The one-hot-encodings satisfy $\sum_{k=1}^{d_j} \XB_{i, j, k} = 1$ for $j=1, \ldots, p$, meaning that the columns of each block sum to $\mathbf{1}$, making $\XB$ not of full rank by construction.
  \item[(P2)] Choosing the number of intervals $d_j$ for binarization of each raw feature $j$ is not an easy task, as too many might lead to overfitting: the number of model-weights increases with each $d_j$, leading to a over-parametrized model.
  \item[(P3)] Some of the raw features $\bX_{\bullet, j} $ might not be relevant for the prediction task, so we want to select raw features from their one-hot encodings, namely induce block-sparsity in $\theta$.
\end{enumerate}
A usual way to deal with (P1) is to impose a linear constraint~\cite{agresti2015foundations} in each block.
In order to do so, let us introduce first $n_{j, k} = | \{ i : x_{i, j} \in I_{j, k} \} |$ and
the vector $n_j = [n_{j, 1} \cdots n_{j, d_j}] \in \N^{d_j}$.
In our penalization term, we impose the linear constraint
\begin{equation}
  \label{eq:linear_constraint}
  n_j^\top \theta_{j, \bul} = \sum_{k=1}^{d_j} n_{j, k} \theta_{j,k} = 0
\end{equation}
for all $j=1, \ldots, p$.
Note that if the $I_{j, k}$ are taken as interquantiles intervals, then for each~$j$, we have that $n_{j, k}$ for $k=1, \ldots, d_j$ are equal and the constraint~\eqref{eq:linear_constraint} becomes
 the standard constraint $\sum_{k=1}^{d_j} \theta_{j,k} = 0$.

The trick to tackle (P2) is to remark that within each block, binary features are ordered.
We use a within block total-variation penalization 
\begin{equation*}
\sum_{j=1}^p \norm{\theta_{j, \bullet}}_{\TV,\hat w_{j,\bullet}}
\end{equation*}
where
\begin{equation}
  \label{eq:tv_no_linear_constraint}
  \norm{\theta_{j,\bullet}}_{\TV,\hat w_{j,\bullet}} = \sum_{k=2}^{d_j} \hat w_{j,k} |\theta_{j, k} - 
  \theta_{j, k-1}|,
\end{equation}
with weights $\hat w_{j, k} > 0$ to be defined later, to keep the number of different values taken by $\theta_{j, \bullet}$ to a minimal level.

\begin{figure}[htb!]
  \centering
  \begin{subfigure}{}
  \includegraphics[width=\columnwidth]{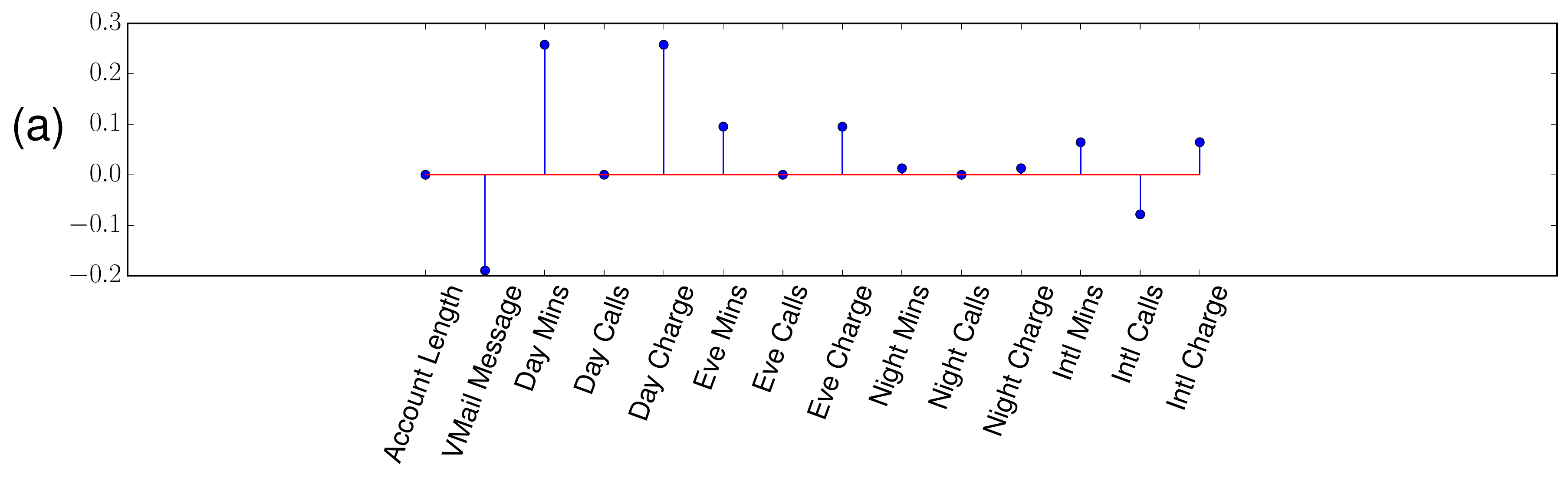}
  \end{subfigure}
  \vspace{-.5cm}
  \begin{subfigure}{}
  \includegraphics[width=\columnwidth]{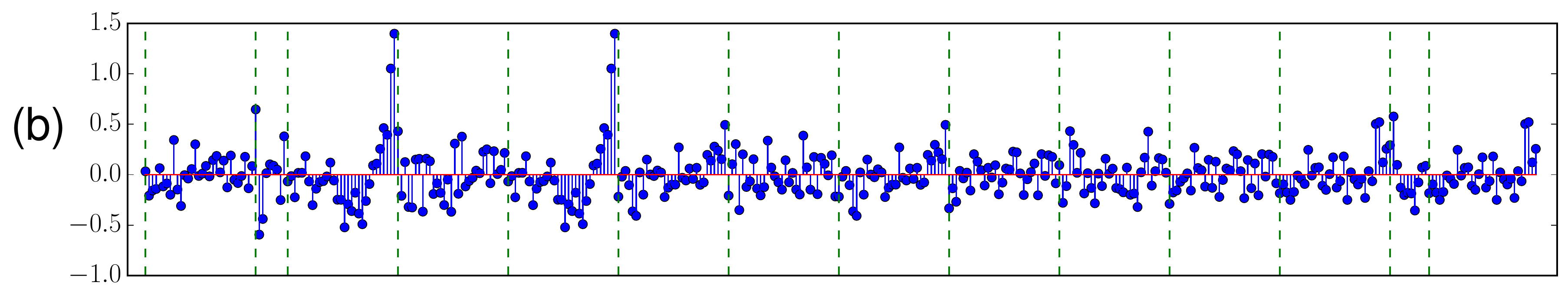}
  \end{subfigure}
  \vspace{-.4cm}
  \begin{subfigure}{}
  \includegraphics[width=\columnwidth]{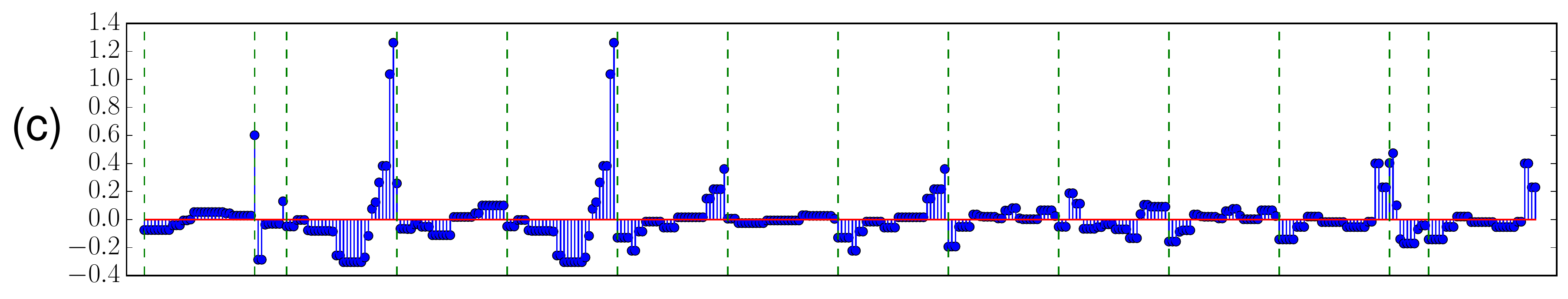} 
  \end{subfigure}
  \vspace{-.4cm}
  \begin{subfigure}{}
  \includegraphics[width=\columnwidth]{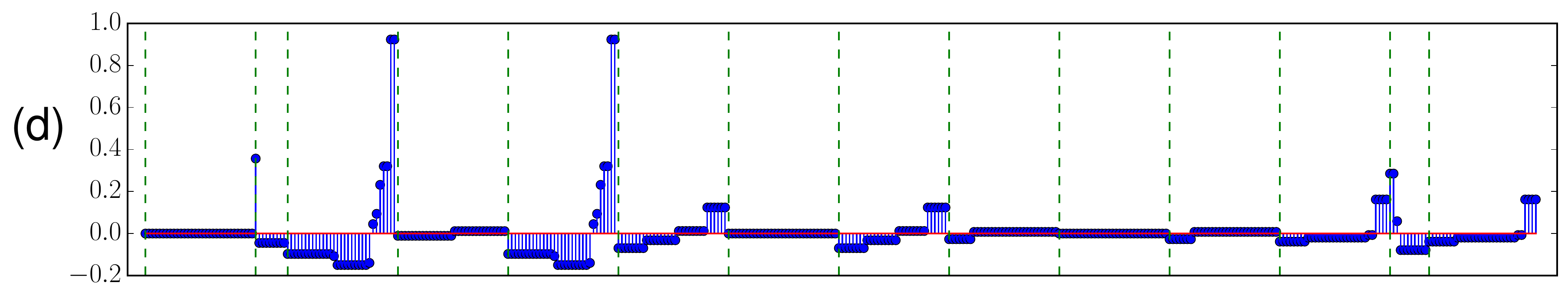}
  \end{subfigure}
\caption{\small Illustration of the binarsity penalization on the ``Churn'' dataset (see Section~\ref{section:btv-experiments} for details) using logistic regression. Figure~(a) shows the model weights learned by the Lasso method on the continuous raw features. Figure~(b) shows the unpenalized weights on the binarized features, where the dotted green lines mark the limits between blocks corresponding to each raw features. Figures~(c) and (d) show the weights with medium and strong binarsity penalization respectively. We observe in (c) that some significant cut-points start to be detected, while in (d) some raw features are completely removed from the model, the same features as those removed in (a).}
\label{figure-binarization-features-agg}  
\end{figure}

Finally, dealing with (P3) is actually a by-product of dealing with (P1) and (P2).
Indeed, if the raw feature $j$ is not-relevant, then $\theta_{j, \bullet}$ should have all entries constant because of the penalization~\eqref{eq:tv_no_linear_constraint}, and in this case all entries are zero, because of~\eqref{eq:linear_constraint}.
We therefore introduce the following penalization, called \emph{binarsity}
\begin{equation}
  \label{eq:binarsity}
   \bina(\theta) = \sum_{j=1}^p \Big( \sum_{k=2}^{d_j} \hat w_{j,k}| 
    \theta_{j, k} -\theta_{j, {k-1}}| + \delta_j(\theta_{j, \bullet}) \Big)
\end{equation}
where the weights $\hat w_{j, k} > 0$ are defined in Section~\ref{sec:theoretical_results} below, and where
\begin{equation}
  \label{eq:def_delta_j}
  \delta_j(u) = 
  \begin{cases}
    0 \quad &\text{ if } \quad n_j^\top u = 0, \\
    \infty &\text{ otherwise}.
  \end{cases}
\end{equation}
We consider the goodness-of-fit~\eqref{eq:gof} penalized by~\eqref{eq:binarsity}, namely
\begin{equation}
\label{model:general_0}
\hat \theta \in \argmin_{\theta \in \R^d } \big\{R_n(\theta) + 
\bina(\theta) \big\}.
\end{equation}
An important fact is that this optimization problem is numerically cheap, as explained in the next paragraph.
Figure~\ref{figure-binarization-features-agg} illustrates the effect of the binarsity penalization with a varying strength on an example. 

\begin{figure*}[htb!]
  \centering
  \includegraphics[width=0.9\textwidth]{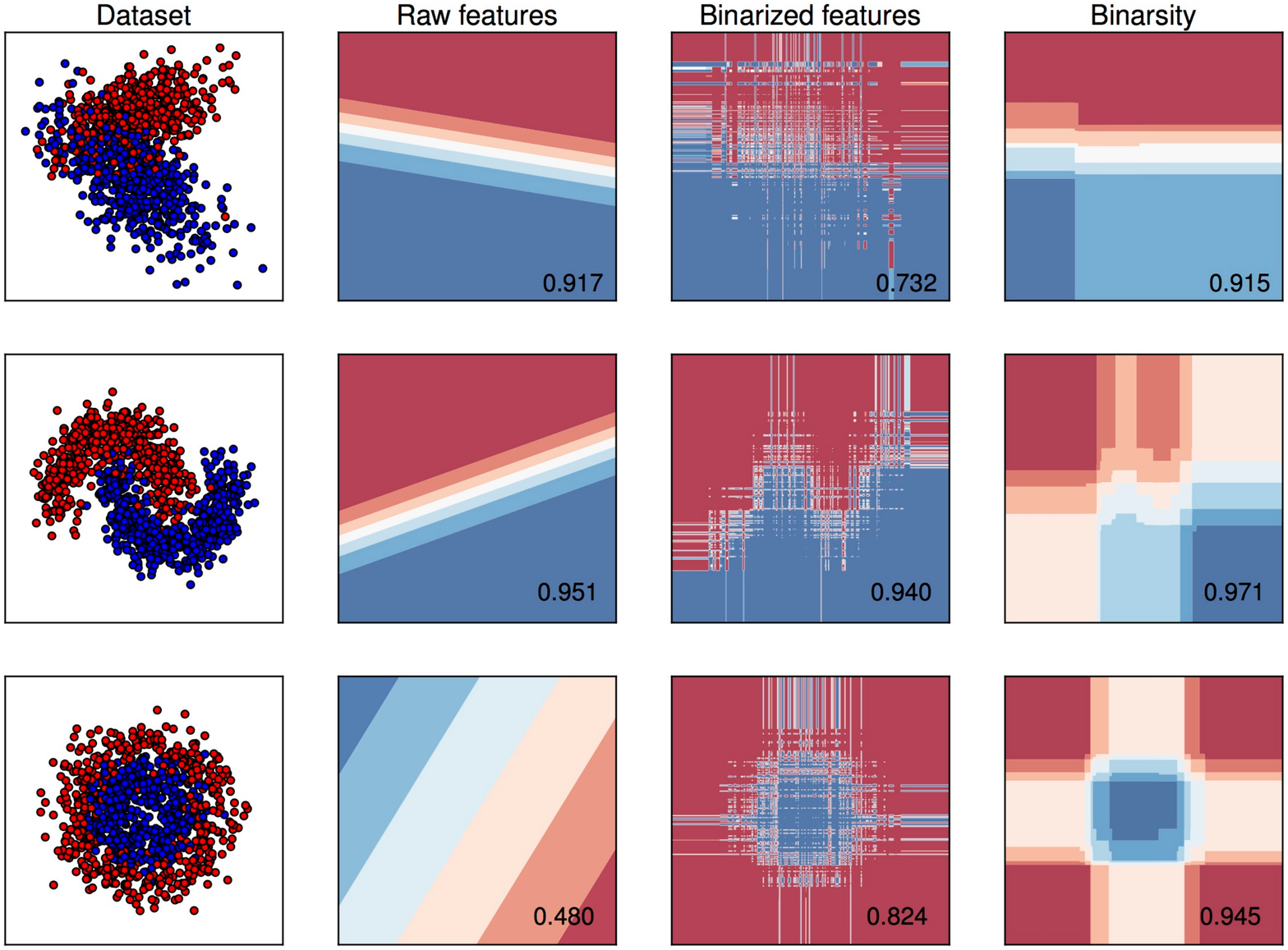}
\caption{\small Illustration of binarsity on 3 simulated toy datasets for binary classification with two classes (blue and red points). We set $n=1000$, $p=2$ and $d_1=d_2=100$. 
In each row, we display the simulated dataset, followed by the decision boundaries for a logistic regression classifier trained on initial raw features, then on binarized features without regularization, and finally on binarized features with binarsity. 
The corresponding testing AUC score is given on the lower right corner of each figure.
Our approach allows to keep an almost linear decision boundary in the first row, while a good decision boundaries are learned on the two other examples, which correspond to non-linearly separable datasets, without apparent overfitting.}
\label{figure-DecisionFunc}
\end{figure*}

In Figure~\ref{figure-DecisionFunc}, we illustrate on a toy example, when $p=2$, the decision boundaries obtained for logistic regression (LR) on raw features, LR on binarized features and LR on binarized features with the binarsity penalization.

\paragraph{Proximal operator of binarsity.}

The proximal operator and proximal algorithms are important tools for non-smooth convex optimization, with important applications in the field  of supervised learning with structured sparsity \cite{bach2012optimization}.
The proximal operator of a proper lower semi-continuous~\cite{BauCom-11} convex function $g : \R^d \rightarrow \R$ is defined by
\begin{equation*}
  \prox_g(v) \in \argmin_{u \in \R^d} \Big\{\frac 12 
  \norm{v- u}_2^2 + g(u) \Big\}.
\end{equation*}
Proximal operators can be interpreted as generalized projections.
Namely, if $g$ is the indicator of a convex set $C \subset \R^d$ given by
\begin{equation*}
  g(u) = \delta_C(u) =
  \begin{cases}
    0 &\text{ if } u \in C, \\
    \infty &\text{ otherwise, }
 \end{cases}
\end{equation*}
then $\prox_g$ is the projection operator onto $C$.
It turns out that the proximal operator of binarsity can be computed very efficiently, using an algorithm~\cite{Cond-13} that we modify in order to include weights $\hat w_{j, k}$. 
It applies in each group the proximal operator of the total-variation since binarsity penalization is block separable, followed by a simple projection onto $\text{span}(n_j)^\perp$ the orthogonal 
of~$\text{span}(n_j)$, see Algorithm~\ref{algorithm-primal-computation} below. 
We refer to Algorithm~\ref{algorithm-weighted-TV-agg} in Section~\ref{appendix:proximal-operator-wTV} for the weighted total-variation proximal operator.
\begin{proposition}
\label{proposition:prox-btv-primal} 
Algorithm~\ref{algorithm-primal-computation} computes the proximal operator of $\bina(\theta)$ given by~\eqref{eq:binarsity}.
\LinesNotNumbered
\begin{algorithm}[htp!]
\SetNlSty{textbf}{}{.}
\DontPrintSemicolon
   \caption{Proximal operator of $\bina(\theta)$, see~\eqref{eq:binarsity}}
   \label{algorithm-primal-computation}
\KwIn{vector $\theta \in \R^d$ and weights $\hat w_{j, k}$ and $n_{j, k}$ for $j=1, \ldots, p$ 
and $k=1, \ldots, d_j$}
\KwOut{vector $\eta = \prox_{\bina}(\theta)$} 
\For{$j=1$ {\bfseries to} $p$}{
$\beta_{j,\bullet}  \gets \prox_{\norm{\theta_{j,\bullet}}_{\TV,\hat w_{j,\bullet}}}(\theta_{j, \bullet})$ (TV-weighted prox in block $j$, see~\eqref{eq:tv_no_linear_constraint}) \\
$\eta_{j,\bullet} \gets  \beta_{j,\bullet} - 
\frac{n_j^\top \beta_{j,\bullet}}{\norm{n_j}_2^2} n_j$ (projection onto $\text{span}(n_j)^\perp$)
}
\textbf{Return:} {$\eta$} 
\end{algorithm} 
\end{proposition}
A proof of Proposition~\ref{proposition:prox-btv-primal} is given in Section~\ref{appendix:proof-of-proposition:prox-btv-primal}.
Algorithm~\ref{algorithm-primal-computation} leads to a very fast numerical routine, see Section~\ref{section:btv-experiments}. 
The next section provides a theoretical analysis of our algorithm with an oracle inequality for the prediction error, together with a convergence rate in the particular case of a sparse additive model.

\section{Theoretical guarantees}
\label{sec:theoretical_results}
 
We now investigate the statistical properties of~\eqref{model:general} where the weights in the binarsity penalization have the form
\begin{equation*}
\hat w_{j,k} = \bigO \Big(\sqrt{\frac{\log d}{n} \ \hat \pi_{j,k}} \Big), \quad 
\text{ with } 
\quad \hat \pi_{j,k} = \frac{\big|\big\{ i=1, \ldots, n: x_{i,j} \in \cup_{k'=k}^{d_j} I_{j, k'}
 \big\}\big|}{n}
\end{equation*}
for all $k \in \{2, \ldots, d_j\}$, see Theorem~\ref{thm:oracle} for a precise definition of $\hat w_{j, k}$.
Note that $\hat \pi_{j,k}$ corresponds to the proportion of ones in the sub-matrix obtained by deleting the first $k$ columns in the $j$-th binarized block matrix $\XB_{\bullet,j}.$
In particular, we have $\hat \pi_{j,k} > 0$ for all $j, k$.
We consider the risk measure defined by
\begin{equation*}\label{eqn:risk}
R(m_{\theta}) =\frac{1}{n} \sum_{i=1}^n \big\{- b'(m^0(x_i)) m_{\theta}(x_i) + b(m_{\theta}(x_i))\big\},
\end{equation*}
which is standard with generalized linear models~\cite{vandegeer2008}.

\subsection{A general oracle inequality}

We aim at evaluating how ``close'' to the minimal possible expected risk  our estimated function $m_{\hat{\theta}}$ with $\hat \theta$ given by~\eqref{model:general} is.
To measure this closeness, we establish a non-asymptotic oracle inequality with a fast rate of convergence considering the excess risk of 
$m_{\hat\theta}$, namely $R(m_{\hat\theta}) - R(m^0)$. 
To derive this inequality, we consider for technical reasons the following problem instead 
of~\eqref{model:general_0}:
\begin{equation}
\label{model:general}
\hat \theta \in \argmin_{\theta \in B_d(\rho)} \big\{R_n(\theta) + 
\bina(\theta) \big\},
\end{equation}
where 
\begin{equation*}
B_d(\rho) = \Big \{\theta \in \R^d: \sum_{j=1}^p\norm{\theta_{j , \bullet}}_\infty \leq \rho \Big\}.
\end{equation*}
This constraint is standard in literature for the proof of oracle inequalities for sparse generalized linear models, see for instance~\cite{vandegeer2008}, and is discussed in details below.

We also impose a restricted eigenvalue assumption on $\XB.$
For all $\theta \in \R^d,$ let $J(\theta)=  [J_1 (\theta),\ldots, J_p (\theta)]$ be the 
concatenation of the support sets relative to the total-variation penalization, that is 
\begin{equation*}
J_j(\theta)= \{k = 2, \ldots, d_j \; : \; \theta_{j,k} \neq \theta_{j,k-1} \}.
\end{equation*}
Similarly, we denote $J^\complement(\theta)= \big[J_1^\complement(\theta),\ldots,J_p^\complement(\theta)\big]$ the complementary of $J(\theta).$ 
The restricted eigenvalue condition is defined as follow.
\begin{assumption}
\label{assumption:RE-XB}
 Let $K =[K_1, \ldots, K_p]$ be a concatenation of index sets such that 
 \begin{equation}
  \label{eq:max_K_Jstar}
  \sum_{j=1}^p |K_j| \leq J^\star,
 \end{equation} 
where $J^\star$ is a positive integer.
 Define 
\begin{equation*}
\normalfont
\kappa (K) \in \inf\limits_{\substack{u \in \mathscr{C}_{\TV,\hat w}(K)}\backslash\{\mathbf{0}\}}\Bigg\{\frac{\norm{\XB u}_2}{\sqrt{n} \norm{u_K}_2}  \Bigg\}
\end{equation*}
with
\begin{equation}
\normalfont
\mathscr{C}_{ \TV, \hat w}(K) \stackrel{}{=} \bigg\{u \in \R^d: \sum_{j=1}^p\norm{(u_{j, \bullet})_{{K_j}^\complement}}_{ \TV, \hat w_{j,\bullet}}
 \leq  2\sum_{j=1}^p \norm{(u_{j, \bullet})_{K_j}}_{ \TV, \hat w_{j,\bullet}}  \bigg\}.
\label{C-AGG}
\end{equation}
We assume that the following condition holds
\begin{equation}
\label{eq:kappa-RE-XB}
\normalfont \kappa (K) > 0
\end{equation}
for any $K$ satisfying~\eqref{eq:max_K_Jstar}.
\end{assumption}
The set $\mathscr{C}_{\TV, \hat w}(K)$ is a cone composed by all vectors with a support ``close'' to $K$.  
Theorem~\ref{thm:oracle} gives a risk bound for the estimator $m_{\hat\theta}$.
\begin{theorem}
\label{thm:oracle} 
Let Assumptions~\ref{ass:glm} and~\ref{assumption:RE-XB} be satisfied. Fix $A >0$ and choose 
\begin{equation}
\label{choice-of-weights-sq-slow-GLM}
 \hat w_{j,k} =  \sqrt{\frac{2U_n\phi(A +\log d)}{n}\,\hat \pi_{j,k}}.
\end{equation}
Then, with probability at least $1 -2e^{-A}$, any $\hat \theta$ given by~\eqref{model:general} satisfies
\begin{align*}
\label{fast-oracle-thm}
R(m_{\hat \theta}) - R(m^0) \leq \inf_{\theta} \Big\{ & 3 (R(m_{\theta}) - R(m^0)) \\
  & \quad + \frac{2560 (C_b(C_n + \rho) + 2)}{L_n \kappa^2(J(\theta))} \; |J(\theta)| \; 
  \max_{j=1, \ldots, p}  \norm{(\hat w_{j,\bullet})_{J_j(\theta)}}_\infty^2 \Big\},
\end{align*}
where the infimum is over the set of vectors $\theta \in B_d(\rho)$ such that $n_j^\top 
\theta_{j, \bullet} = 0$ for all $j=1, \ldots, p$ and such that $|J(\theta)| \leq J^*$.
\end{theorem}

The proof of Theorem~\ref{thm:oracle} is given in Section~\ref{proof-fast-oracle-ineq-bina} below.
Note that the ``variance'' term or ``complexity'' term in the oracle inequality satisfies
\begin{equation}
  \label{complex-term-thm1}
  |J(\theta)| \max_{j=1, \ldots, p}\norm{(\hat w_{j,\bullet})_{J_j(\theta)}}_\infty^2 
  \leq 2 U_n\phi \frac{|J(\theta)|(A + \log d)}{n}.
\end{equation} 
The value $|J(\theta)|$ characterizes the sparsity of the vector $\theta$, given by 
\begin{equation*}
  |J(\theta)| = \sum_{j=1}^p |J_j(\theta)| = 
  \sum_{j=1}^p | \{k = 1, \ldots, d_j : \theta_{j,k} \neq \theta_{j,k-1} \}|.
\end{equation*}
It counts the number of non-equal consecutive values of $\theta$. 
If $\theta$ is block-sparse, namely whenever $|\cJ(\theta)| \ll p$ where $\cJ(\theta) = \{ j = 1, \ldots, p : \theta_{j, \bul} \neq 0_{d_j} \}$ (meaning that few raw features are useful for prediction), then 
$|J(\theta)| \leq |\cJ(\theta)| \max_{j \in \cJ(\theta)} |J_j(\theta)|$, which means that $|J(\theta)|$ is controlled by the block sparsity $|\cJ(\theta)|$.

The oracle inequality from Theorem~\ref{thm:oracle} is stated uniformly for vectors $\theta \in B_d(\rho)$ satisfying $n_j^\top \theta_{j, \bullet} = 0$ for all $j=1, \ldots, p$ and $|J(\theta)| \leq J^*$.
Writing this oracle inequality under the assumption $|J(\theta)| \leq J^*$ meets the standard way of stating sparse oracle inequalities, see e.g.~\cite{BuhVan-11}.
Note that $J^*$ is introduced in Assumption~\ref{assumption:RE-XB} and corresponds to a maximal sparsity for which the matrix $\bX^B$ satisfies the restricted eigenvalue assumption.
Also, the oracle inequality stated in Theorem~\ref{thm:oracle} stands for vectors such that $n_j^\top \theta_{j, \bullet} = 0$, which is natural since the binarsity penalization imposes these extra linear constraints.

The assumption that $\theta \in B_d(\rho)$ is a technical one, that allows to establish a connection, via the notion of self-concordance, see~\cite{bach2010selfconcordance}, between the empirical squared $\ell_2$-norm and the empirical Kullback divergence (see Lemma~\ref{lemma-connection-L2-KL} in Section~\ref{proof-fast-oracle-ineq-bina}).
It corresponds to a technical constraint which is commonly used in literature for the proof of oracle inequalities for sparse generalized linear models, see for instance~\cite{vandegeer2008}, a recent contribution for the particular case of Poisson regression being~\cite{ivanoff2016adaptive}.
Also, note that
\begin{equation}
  \label{lemma:control_inner_ball}
  \max_{i=1,\ldots,n} | \inr{x_i^B, \theta} | \leq \sum_{j=1}^p 
  \norm{\theta_{j , \bullet}}_\infty \leq |\cJ(\theta)|
  \times \norm{\theta}_\infty,
\end{equation}
where $\norm{\theta}_\infty = \max_{j=1, \ldots, p} \norm{\theta_{j, \bul}}_\infty$.
The first inequality in~\eqref{lemma:control_inner_ball} comes from the fact that the entries of $\XB$ are in $\{0, 1\}$, and it entails that $ \max_{i=1,\ldots,n} | \inr{x_i^B, \theta} | \leq \rho$ whenever $\theta \in B_d(\rho)$.
The second inequality in~\eqref{lemma:control_inner_ball} entails that $\rho$ can be upper bounded by $|\cJ(\theta)| \times \norm{\theta}_\infty$, and therefore the constraint $\theta \in B_d(\rho)$ becomes only a box constraint on $\theta$, which depends on the dimensionality of the features through $|\cJ(\theta)|$ only.
The fact that the procedure depends on $\rho$, and that the oracle inequality stated in Theorem~\ref{thm:oracle} depends linearly on $\rho$ is commonly found in literature about sparse generalized linear models, see~\cite{vandegeer2008,bach2010selfconcordance,ivanoff2016adaptive}.
However, the constraint $B_d(\rho)$ is a technicality which is not used in the numerical experiments provided in Section~\ref{section:btv-experiments} below.

In the next Section, we exhibit a consequence of Theorem~\ref{thm:oracle}, whenever one considers the Gaussian case (least-squares loss) and where $m^0$ has a sparse additive structure defined below.
This structure allows to control the bias term from Theorem~\ref{thm:oracle} and to exhibit a convergence rate.

\subsection{Sparse linear additive regression} 
\label{sub:sparse_linear_additive_regression}

Theorem~\ref{thm:oracle} allows to study a particular case, namely an additive model, see e.g.~\cite{hastie1990generalized,horowitz2006optimal} and in particular a sparse additive linear model, which is of particular interest in high-dimensional statistics, see~\cite{meier2009high,ravikumar2009sparse,BuhVan-11}.
We prove in Theorem~\ref{thm:additive} below that our procedure matches the convergence rates previously known from literature.
In this setting, we work under the following assumptions.

\begin{assumption}
  \label{ass:additive-model}
  We assume to simplify that $x_i \in [0, 1]^d$ for all $i=1, \ldots, n$.
  We consider the Gaussian setting with the least-squares loss, 
  namely $\ell(y, y') = \frac 12 (y - y')^2$, $b(y) = \frac 12 y^2$ and $\phi = \sigma^2$ (noise variance) in Equation~\eqref{distribut-glm}, with $L_n = U_n = 1$, $C_b = 0$ in 
  Assumption~\ref{ass:glm}.
  Moreover, we assume that $m^0$ has the following sparse additive structure
  \begin{equation*}
    m^0(x) = \sum_{j \in \cJ_*} m_j^0(x_j)
  \end{equation*}
  for $x = [x_1 \cdots x_p] \in \R^p$, where $m_j^0 : \R \rightarrow \R$ are $L$-Lipschitz functions, namely satisfying  $|m_j^0(z) - m_j^0(z')| \leq L |z - z'|$ for any $z, z' \in \R$, and where $\cJ_* \subset \{ 1, \ldots, p \}$ is a set of active features (sparsity means that $|\cJ_*| \ll p$). 
  Also, we assume the following identifiability condition
  \begin{equation*}
    \sum_{i=1}^n m_j^0(x_{i, j}) = 0
  \end{equation*}
  for all $j=1, \ldots, p$.
\end{assumption}

Assumption~\ref{ass:additive-model} contains identifiability and smoothness requirements that are standard when studying additive models, see e.g.~\cite{meier2009high}.
We restrict the functions $m_j^0$ to be Lipschitz and not smoother, since our procedure produces a piecewise constant decision function with respect to each $j$, that can approximate optimally only Lipschitz functions.
For more regular functions, our procedure would lead to suboptimal rates, see also the discussion below the statement of Theorem~\ref{thm:additive}.

\begin{theorem}
  \label{thm:additive}
  Consider procedure~\eqref{model:general_0} with $d_j = D$, where $D$ is the integer part of $n^{1/3}$, and $I_{j, 1} = [0, \frac{1}{D}]$, $I_{j, k} = (\frac{k-1}{D}, \frac{k}{D}]$ for all $k=2, \ldots, D$ and $j=1, \ldots, p$, and keep the weights $\hat w_{j, k}$ the same as in Theorem~\ref{thm:oracle}.
  Introduce also $\theta_{j, k}^* = \sum_{i=1}^n m_j^0(x_{i, j}) \ind{I_k}(x_{i, j}) / \sum_{i=1}^n \ind{I_k}(x_{i, j})$ for $j \in \cJ_*$ and $\theta_{j, \bul}^* = \boldsymbol 0_D$ for $j \notin \cJ_*$.
  Then, under Assumption~\ref{assumption:RE-XB} with $J^* = J(\theta^*)$ and Assumption~\ref{ass:additive-model}, we have
  \begin{equation*}
  \norm{m_{\hat \theta} - m^0}_n^2 \leq \Big(3 L^2 |\cJ_*| + \frac{5120 M_n \sigma^2 
  (A + \log(p n^{1/3} M_n)}{\kappa^2(J(\theta^*))} \Big) \frac{|\cJ_*|}{n^{2/3}},
  \end{equation*}
  where $M_n = \max_{j=1, \ldots, p} \max_{i=1, \ldots, n} |m_j^0(x_{i, j})|.$
\end{theorem}
The proof of Theorem~\ref{thm:additive} is given in Section~\ref{sub:proof_of_theorem_thm:additive} below.
It is an easy consequence of Theorem~\ref{thm:oracle} under the sparse additive model assumption.
It uses Assumption~\ref{assumption:RE-XB} with $J^* = J(\theta^*)$, since $\theta_{j, \bul}^*$ is the minimizer of the bias for each $j \in \cJ_*$, see the proof of Theorem~\ref{thm:additive} for details.

The rate of convergence is, up to constants and logarithmic terms, of order $|\cJ_*|^2 n^{-2/3}.$
Recalling that we work under a Lipschitz assumption, namely H\"older smoothness of order $1$, the scaling of this rate w.r.t. to $n$ is $n^{-2 r / (2 r + 1)}$ with $r=1$, which matches the one-dimensional minimax rate.
This rate matches the one obtained in~\cite{BuhVan-11}, see Chapter~8 p.~272, where the rate $|\cJ_*|^2 n^{-2r / (2r + 1)} = |\cJ_*|^2 n^{-4 / 5}$ is derived under a $C^2$ smoothness assumption, namely $r = 2$.
Hence, Theorem~\ref{thm:additive} shows that, in the particular case of a sparse additive model, our procedure matches in terms of convergence rate the state of the art.
Further improvements could consider more general smoothness (beyond Lipschitz) and adaptation with respect to the regularity, at the cost of a more complicated procedure which is beyond the scope of this paper.

\section{Numerical experiments}
\label{section:btv-experiments}

\begin{figure}[htp!]
\centering
\includegraphics[width=.75\columnwidth]{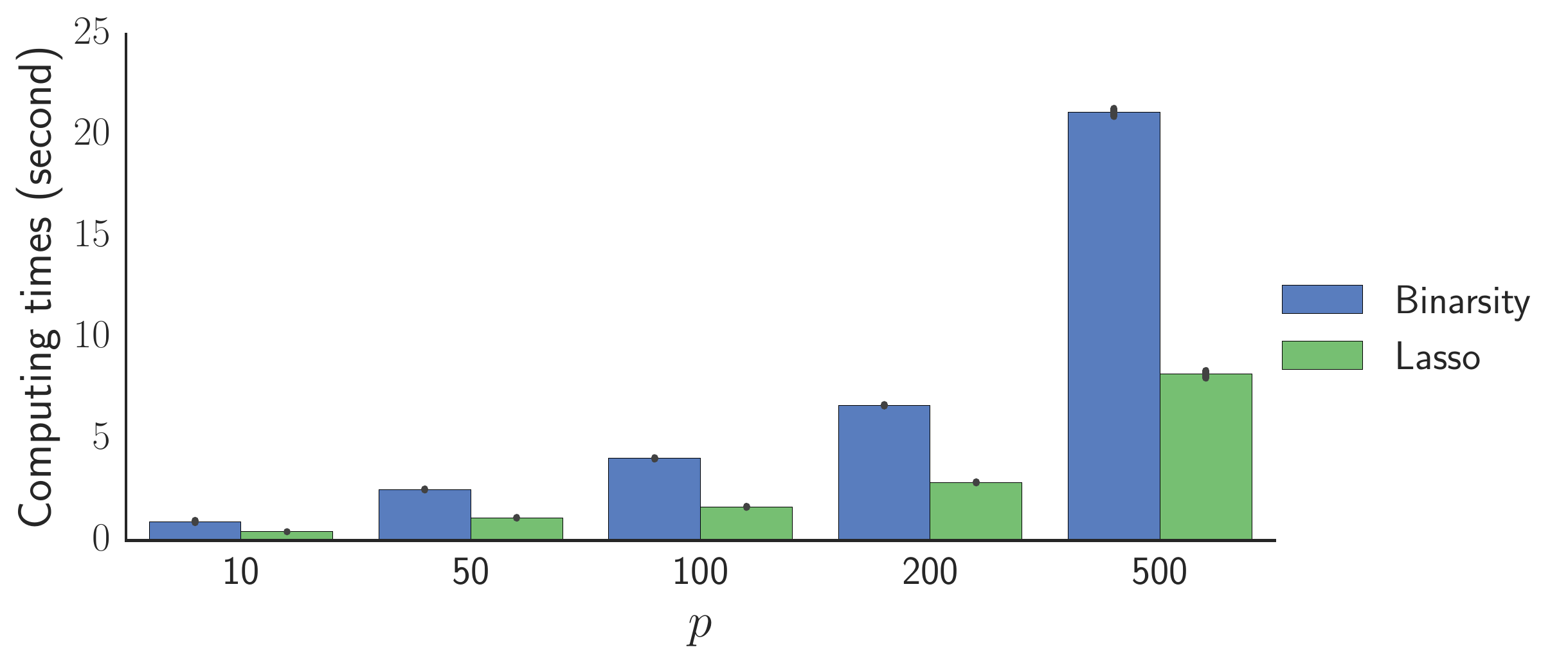}\hfill
\caption{\small Average computing time in second (with the black lines representing $\pm$ the standard deviation) obtained on 100 simulated datasets for training a logistic model with binarsity VS Lasso penalization, both trained on $\XB$ with $d_j=10$ for all $j \in 1, \ldots, p$. Features are Gaussian with a Toeplitz covariance matrix with correlation $0.5$ and $n=10000$. Note that the computing time ratio between the two methods stays roughly constant and equal to $2$.}
\label{fig:computing-times-simu}
\end{figure}

In this section, we first illustrate the fact that the binarsity penalization is roughly only two times slower than basic $\ell_1$-penalization, see the timings in Figure~\ref{fig:computing-times-simu}. We then compare binarsity to a large number of baselines, see Table~\ref{table:baselines}, using 9 classical binary classification datasets obtained from the UCI Machine Learning Repository~\cite{Lichman:2013}, see Table~\ref{table:datasets}.

\begin{table}[htp!]
\centering
\small
\begin{tabular}{cccc}
\toprule
 Name & Description & Reference \\
 \midrule
 Lasso & Logistic regression (LR) with $\ell_1$ penalization & \cite{tibshirani1996regression}\\
 Group L1 & LR with group $\ell_1$ penalization & \cite{meier2008group} \\
 Group TV & LR with group total-variation penalization  & \\
 SVM & Support vector machine with radial basis kernel & \cite{scholkopf2002learning} \\
 GAM & Generalized additive model & \cite{hastie1990generalized} \\
 RF & Random forest classifier & \cite{breiman2001random} \\
 GB & Gradient boosting & \cite{friedman2002stochastic} \\
 \bottomrule
\end{tabular}
\caption{\small Baselines considered in our experiments. Note that Group L1 and Group TV are considered on binarized features.}
\label{table:baselines}
\end{table}

\begin{table}[htb!]
\centering
\small
\begin{tabular}{cccc}
\toprule
 Dataset & \#Samples & \#Features & Reference \\
 \midrule
 Ionosphere & 351 & 34 & \cite{sigillito1989classification}\\
 Churn & 3333 & 21 & \cite{Lichman:2013} \\
 Default of credit card & 30000 & 24 & \cite{yeh2009comparisons} \\
  Adult & 32561 & 14 & \cite{kohavi1996scaling} \\
 Bank marketing & 45211 & 17 & \cite{moro2014data} \\
 Covertype & 550088 & 10 & \cite{blackard1999comparative} \\
 SUSY & 5000000 & 18 & \cite{baldi2014searching} \\
 HEPMASS & 10500000 & 28 & \cite{baldi2016parameterized} \\
 HIGGS & 11000000 & 24 & \cite{baldi2014searching} \\
 \bottomrule
\end{tabular}
\caption{\small Basic informations about the 9 considered datasets.}
\label{table:datasets}
\end{table}

\begin{figure}[htp!]
  \centering
\includegraphics[width=0.8\textwidth]{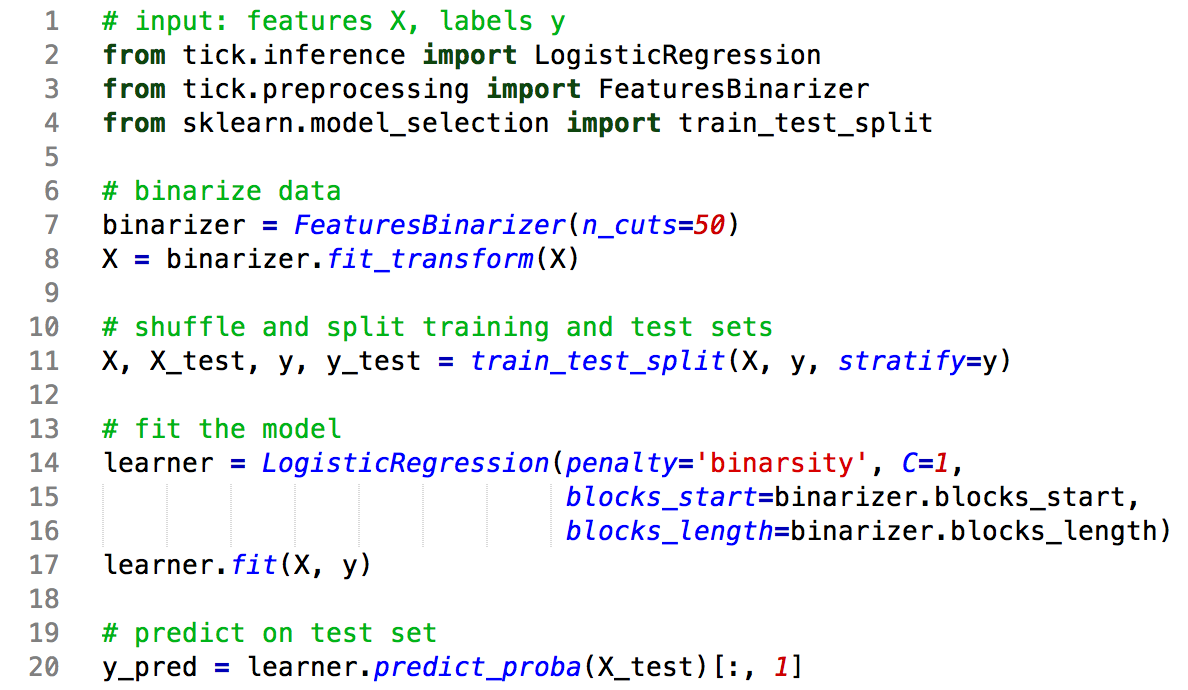}
  \caption{\small Sample python code for the use of binarsity with logistic regression in the \texttt{tick} library, with the use of the \texttt{FeaturesBinarizer} transformer for features binarization.}
  \label{fig:code}
\end{figure}

For each method, we randomly split all datasets into a training and a test set (30\% for testing), and all hyper-parameters are tuned on the training set using $V$-fold cross-validation with
 $V = 10$.
For support vector machine with radial basis kernel (SVM), random forests (RF) and gradient boosting (GB), we use the reference implementations from the \texttt{scikit-learn} library~\cite{scikit-learn}, and we use the \texttt{LogisticGAM} procedure from the~\texttt{pygam} library\footnote{\url{https://github.com/dswah/pyGAM}} for the GAM baseline.
The binarsity penalization is proposed in the~\texttt{tick} library~\cite{2017arXiv170703003B}, we provide sample code for its use in Figure~\ref{fig:code}.
Logistic regression with no penalization or ridge penalization gave similar or lower scores for all considered datasets, and are therefore not reported in our experiments. 

\begin{figure*}[htp!]
\centering
\includegraphics[width=\textwidth]{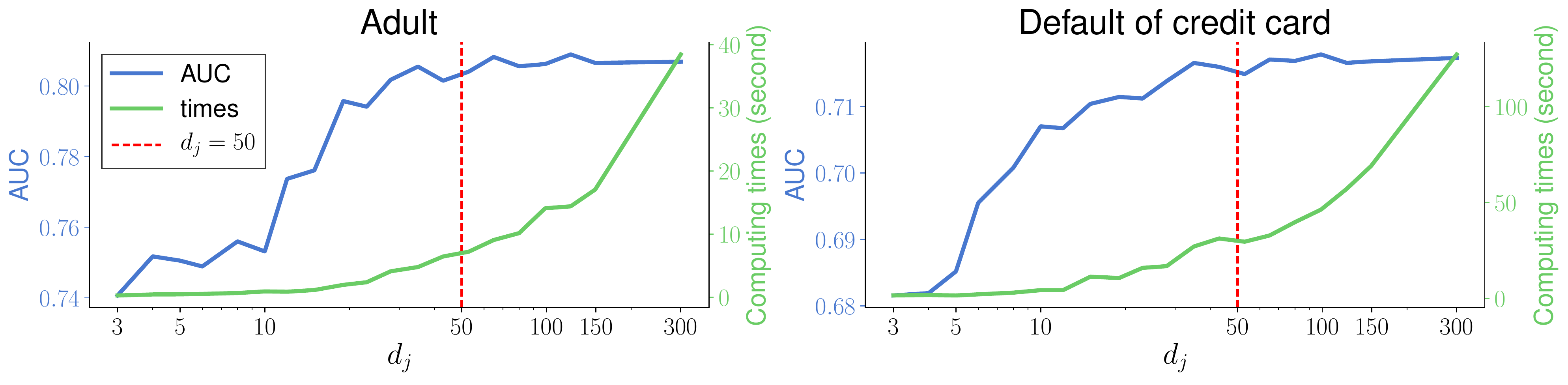}
\caption{\small Impact of the number of bins used in each block ($d_j$) on the classification performance (measured by AUC) and on the training time using the ``Adult'' and ``Default of credit card'' datasets. All $d_j$ are equal for $j = 1, \ldots, p$, and we consider in all cases the best hyper-parameters selected after cross validation. We observe that past $d_j=50$ bins, performance is roughly constant, while training time strongly increases.}
\label{fig:discretization-impact}
\end{figure*}

The binarsity penalization does not require a careful tuning of $d_j$ (number of bins for the one-hot encoding of raw feature $j$).
Indeed, past a large enough value, increasing $d_j$ even  further barely changes the results since the cut-points selected by the penalization do not change anymore.
This is illustrated in Figure~\ref{fig:discretization-impact}, where we observe that past $50$ bins, increasing $d_j$ even further does not affect the performance, and only leads to an increase of the training time.
In all our experiments, we therefore fix $d_j=50$ for $j = 1, \ldots, p$.

\begin{figure*}[!htp]
\centering
\includegraphics[width=.32\textwidth]{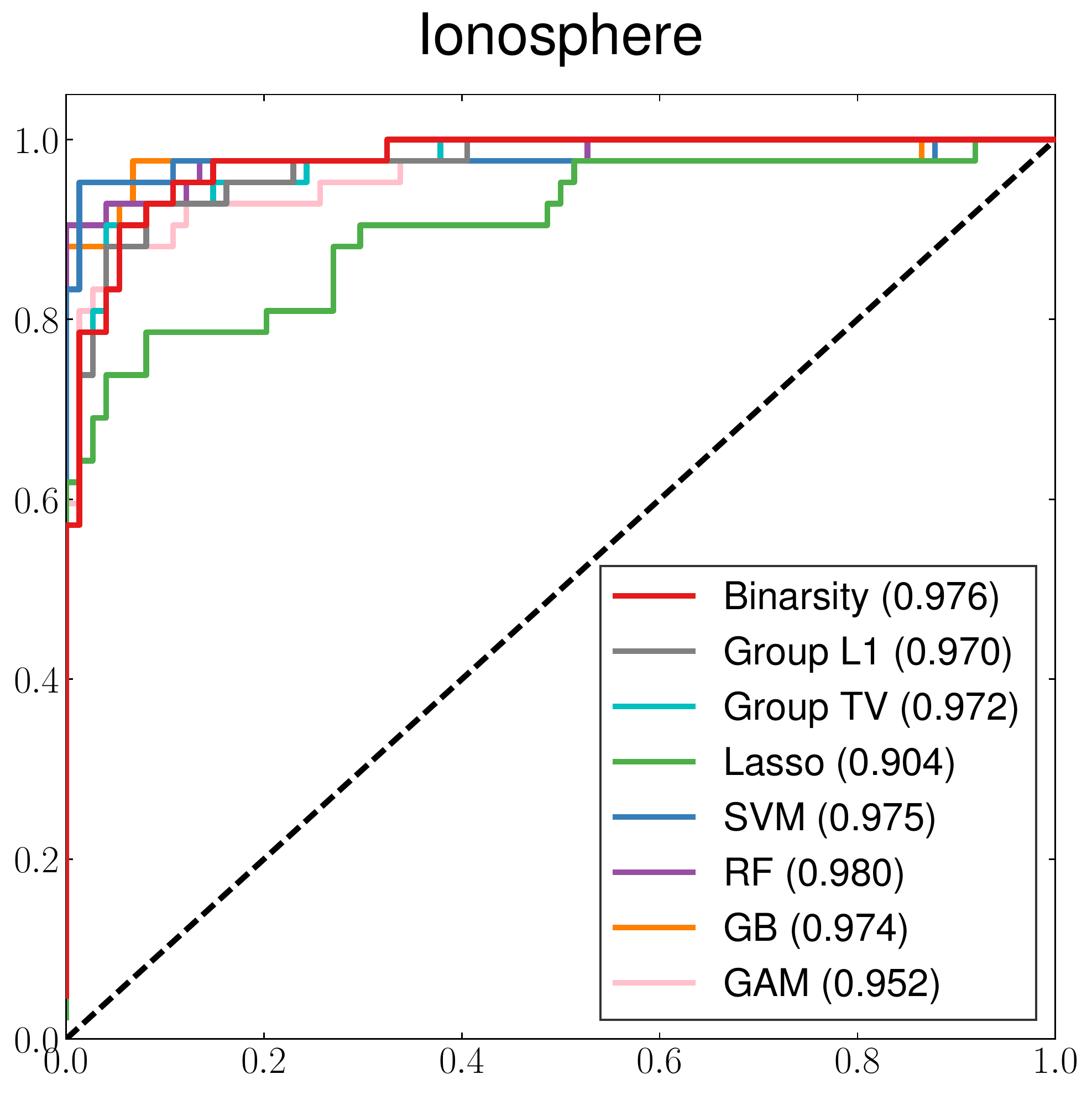}\hfill
\includegraphics[width=.32\textwidth]{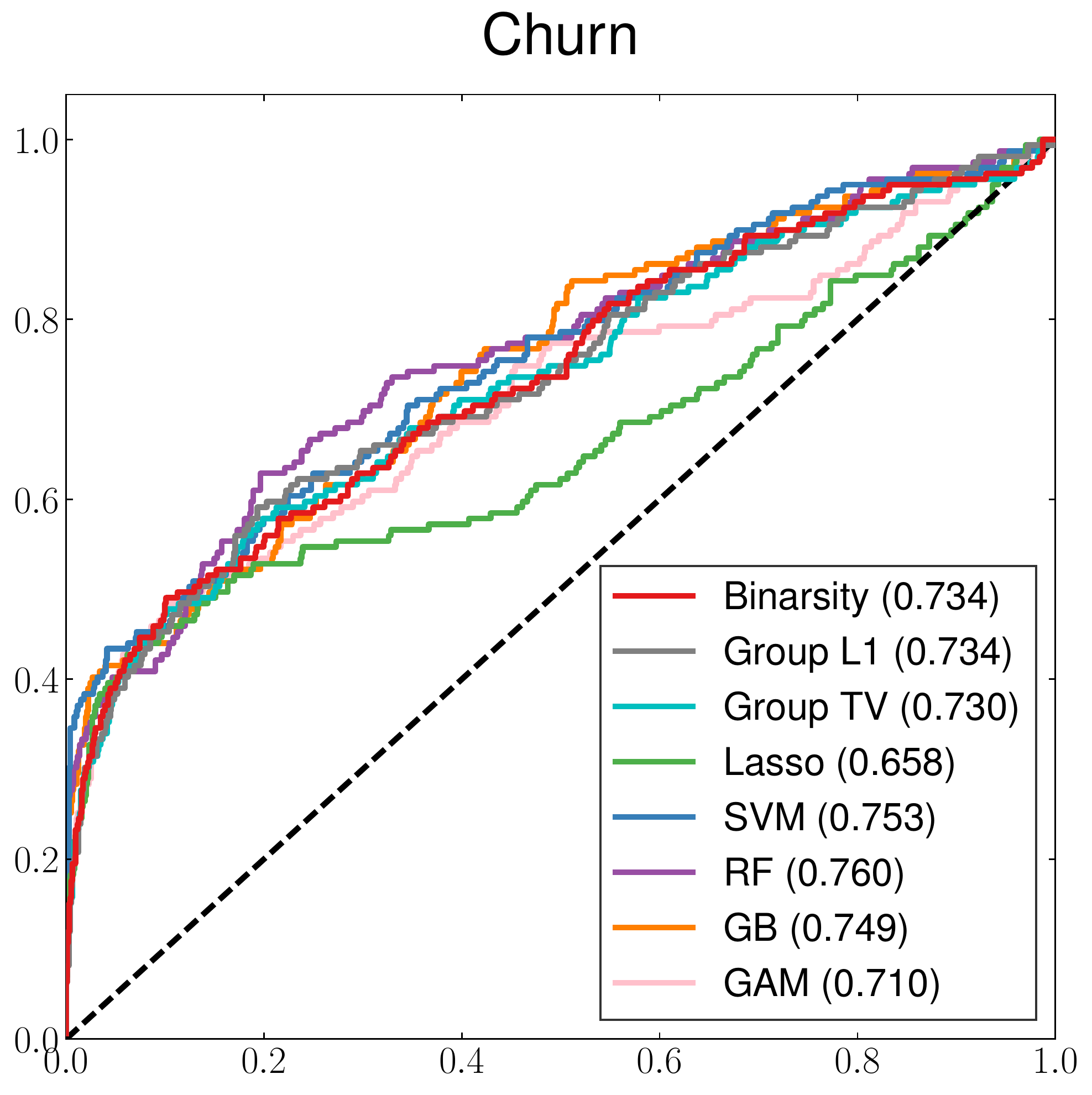}\hfill
\includegraphics[width=.32\textwidth]{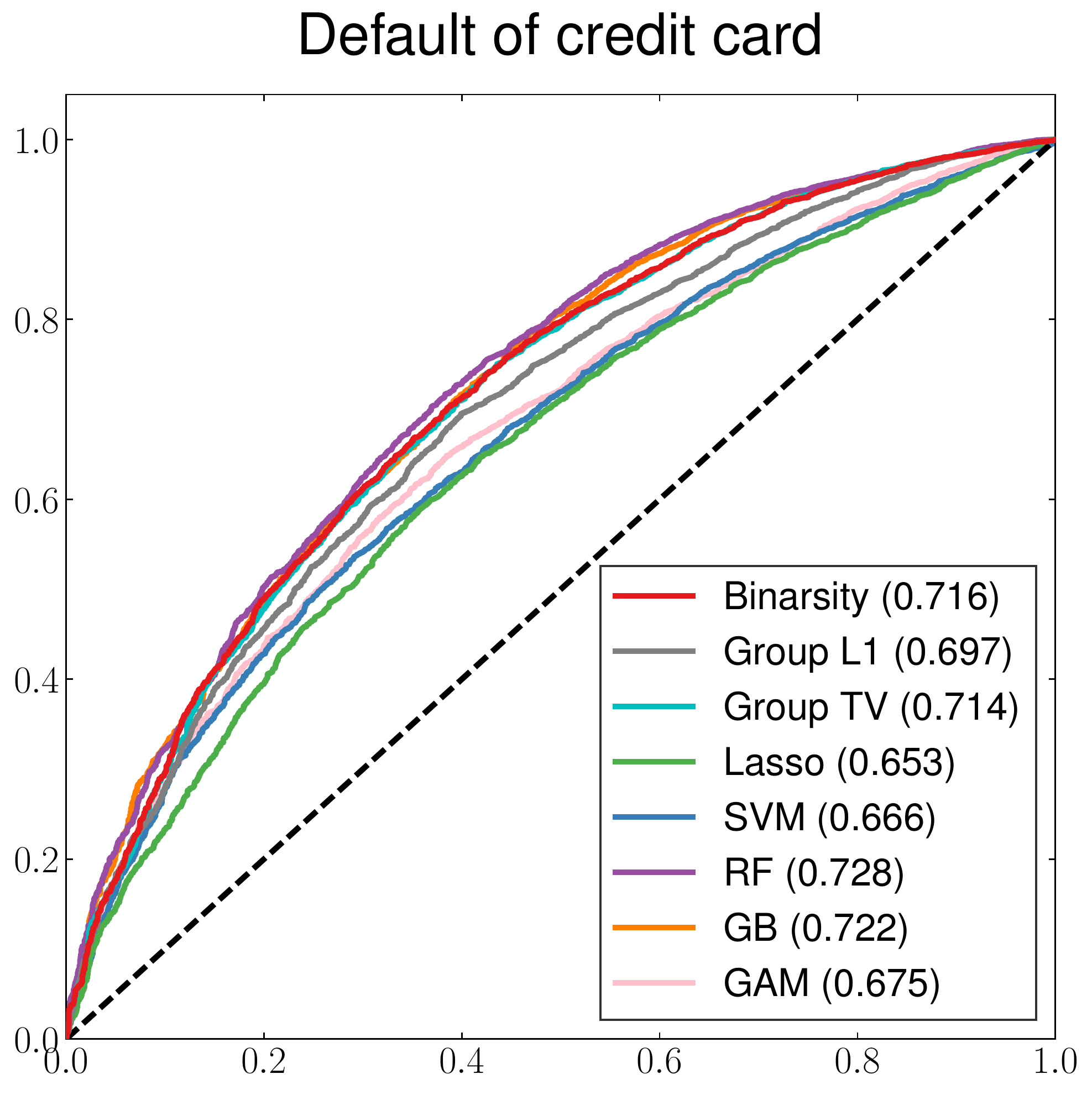}

\includegraphics[width=.32\textwidth]{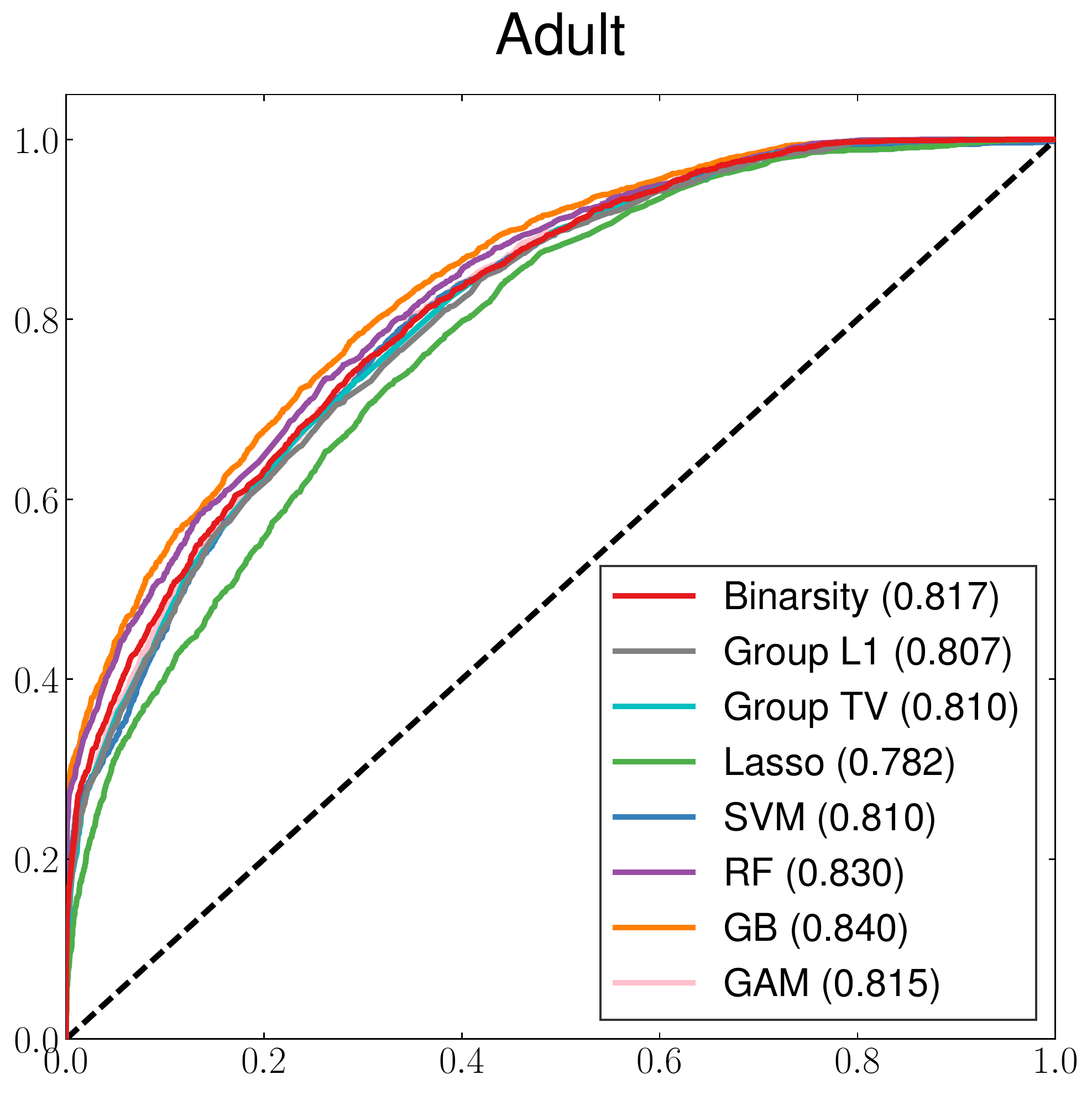}\hfill
\includegraphics[width=.32\textwidth]{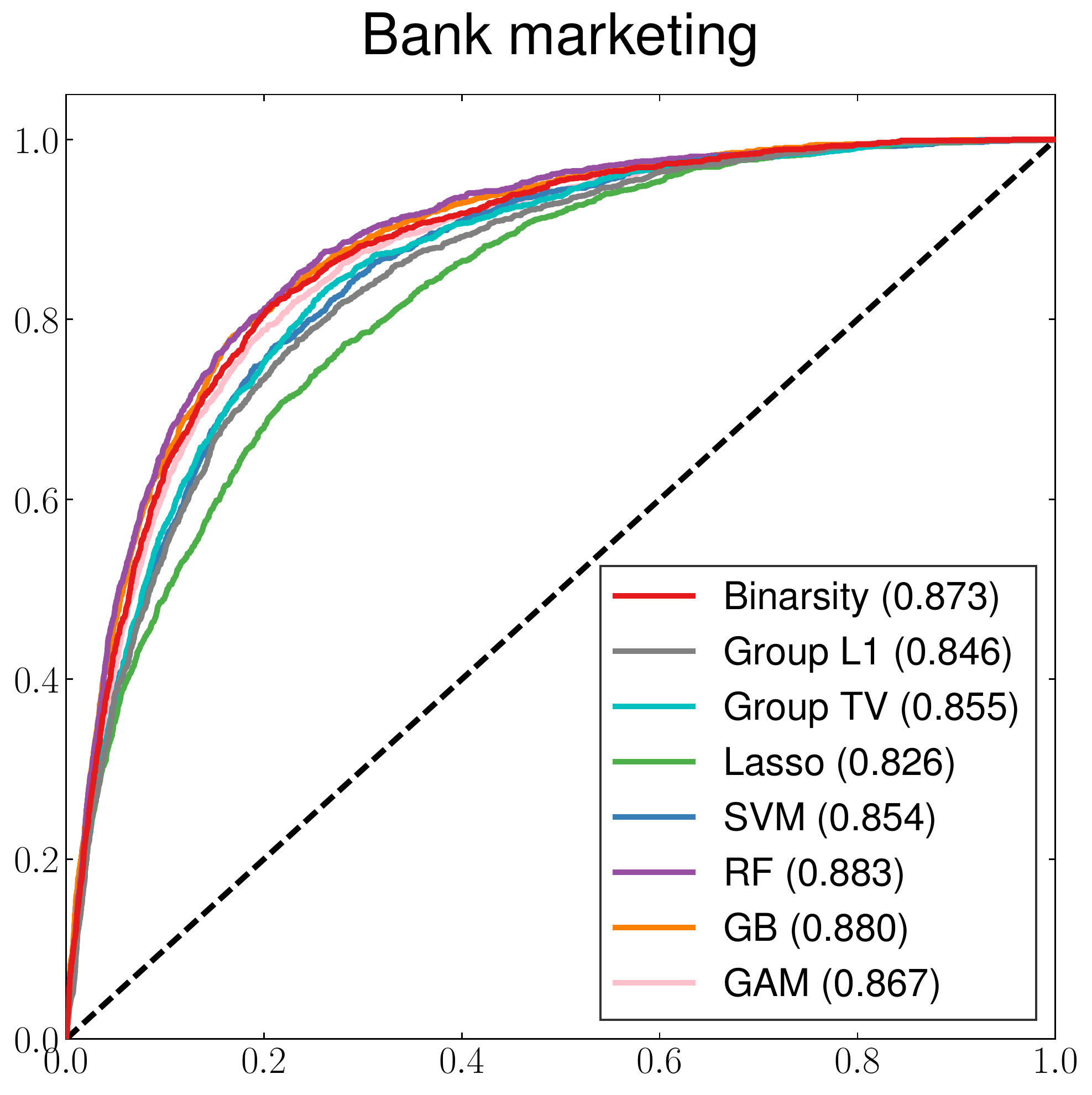}\hfill
\includegraphics[width=.32\textwidth]{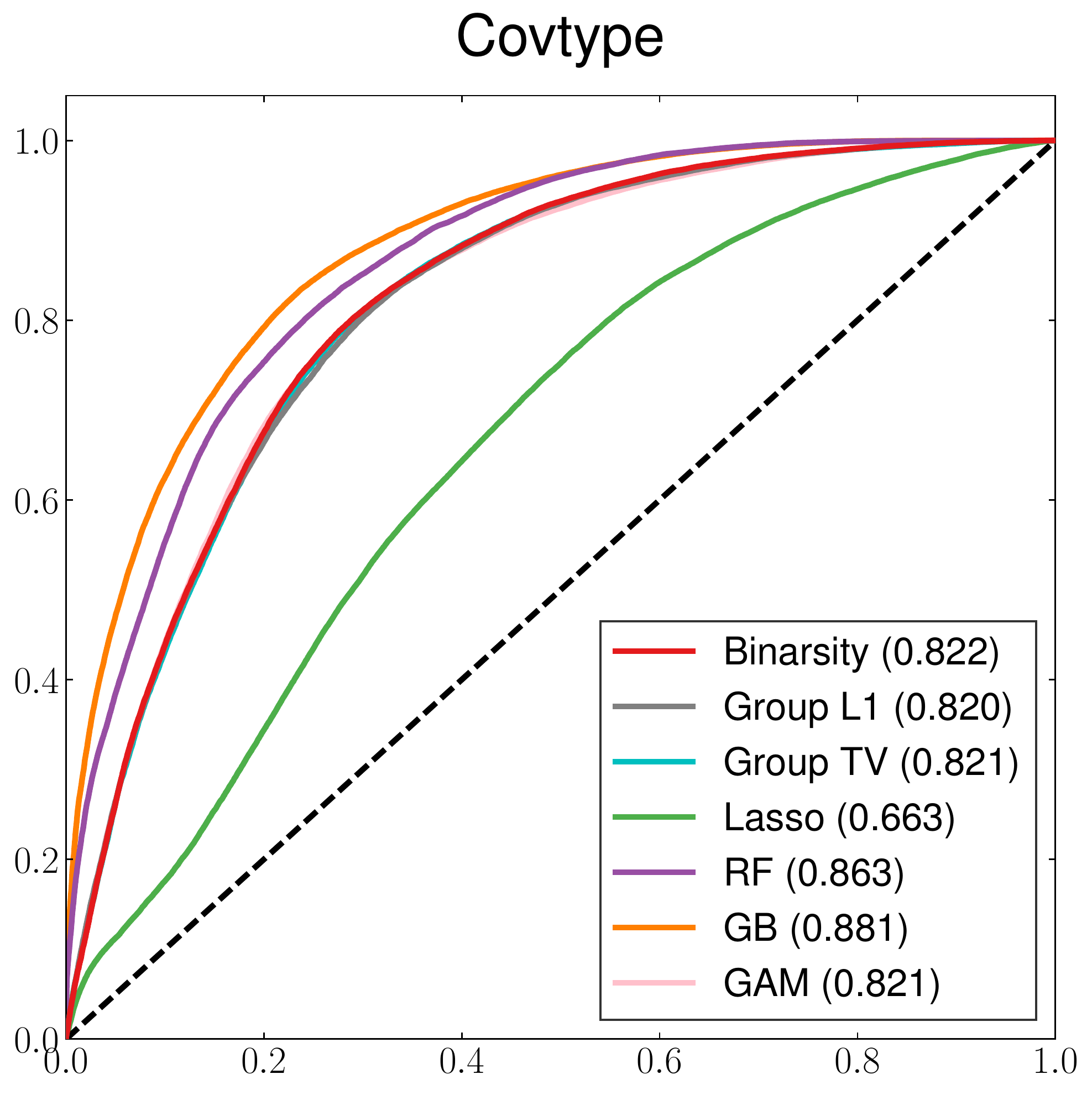}

\includegraphics[width=.32\textwidth]{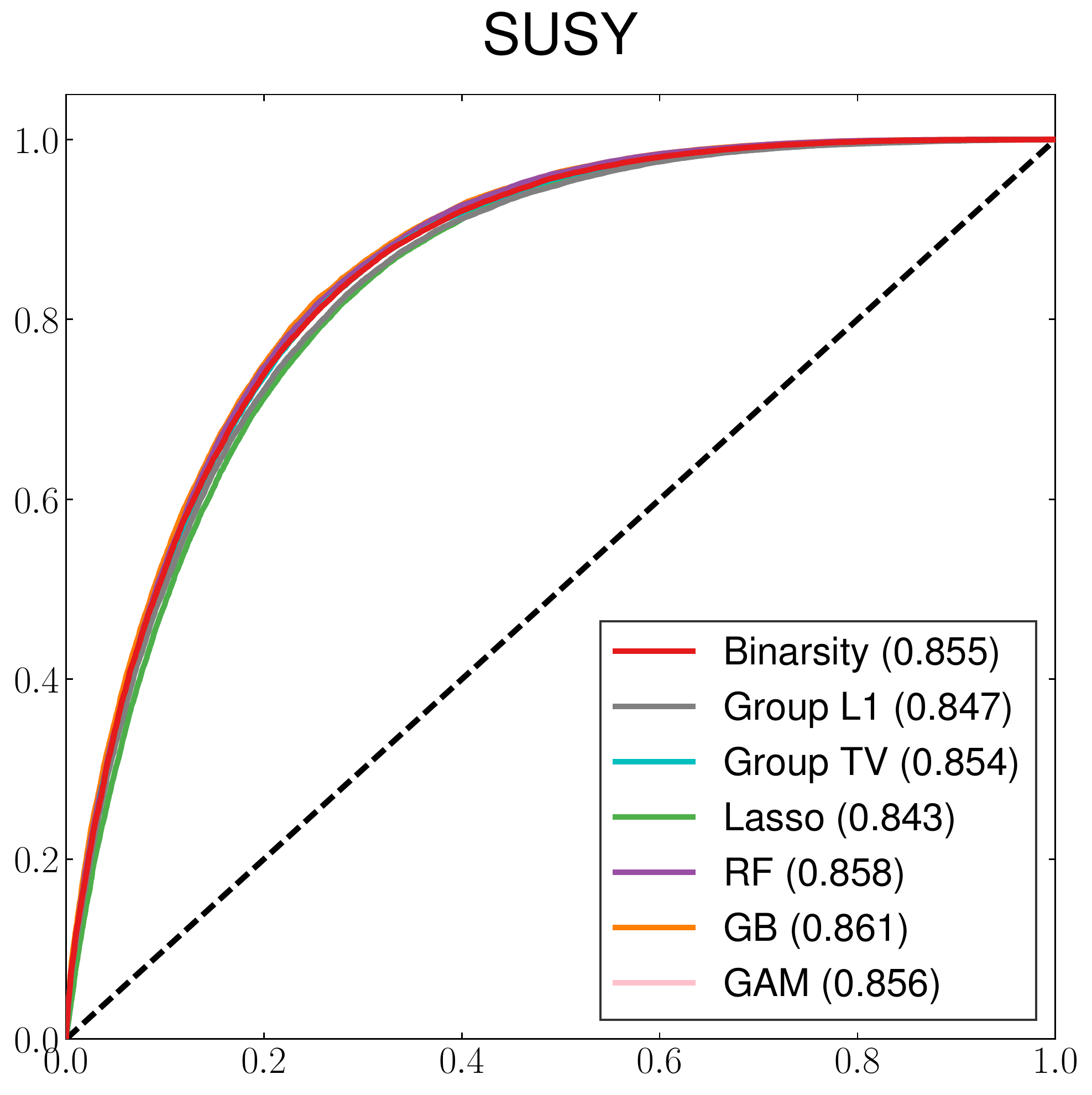}\hfill
\includegraphics[width=.32\textwidth]{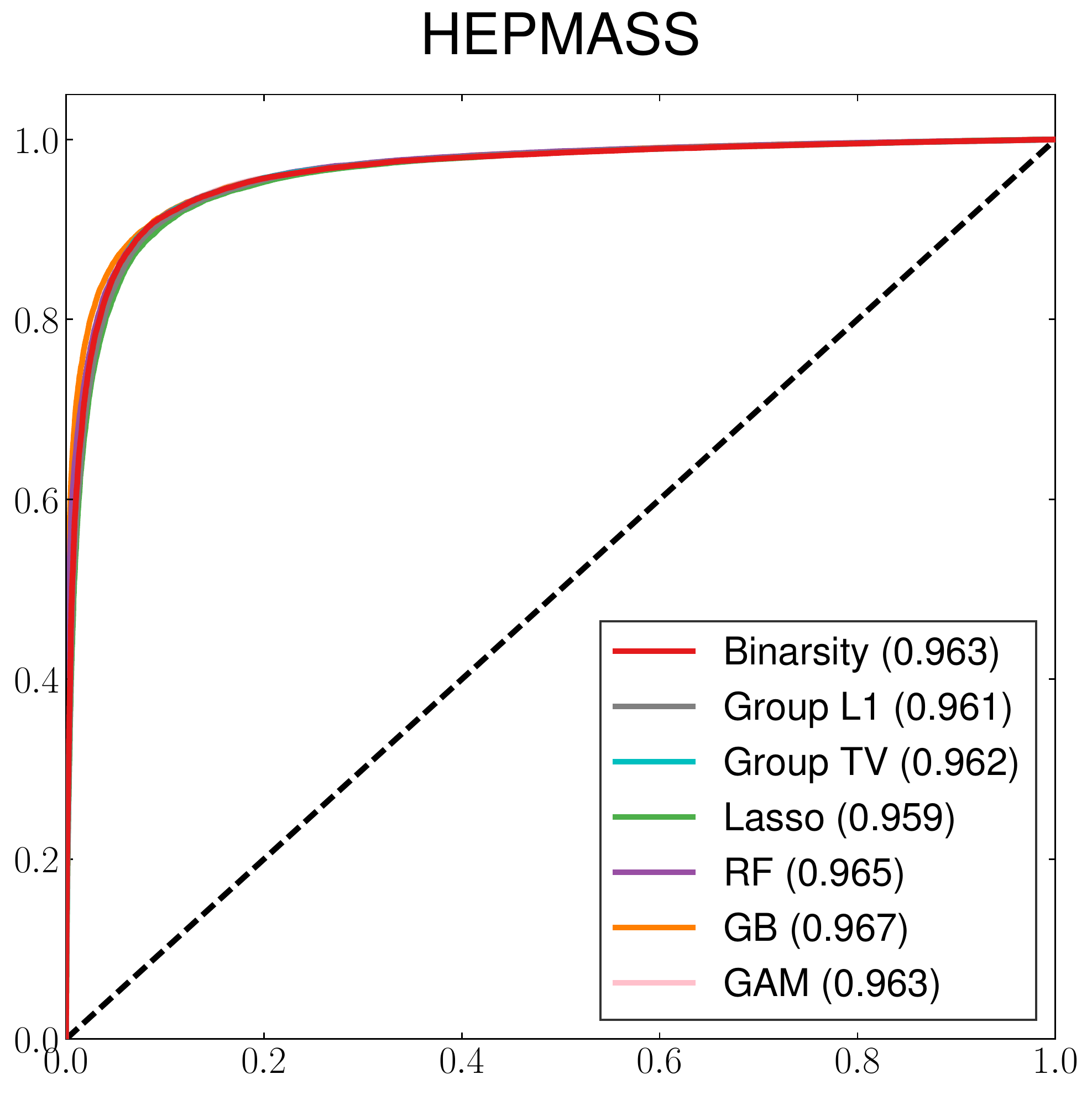}\hfill
\includegraphics[width=.32\textwidth]{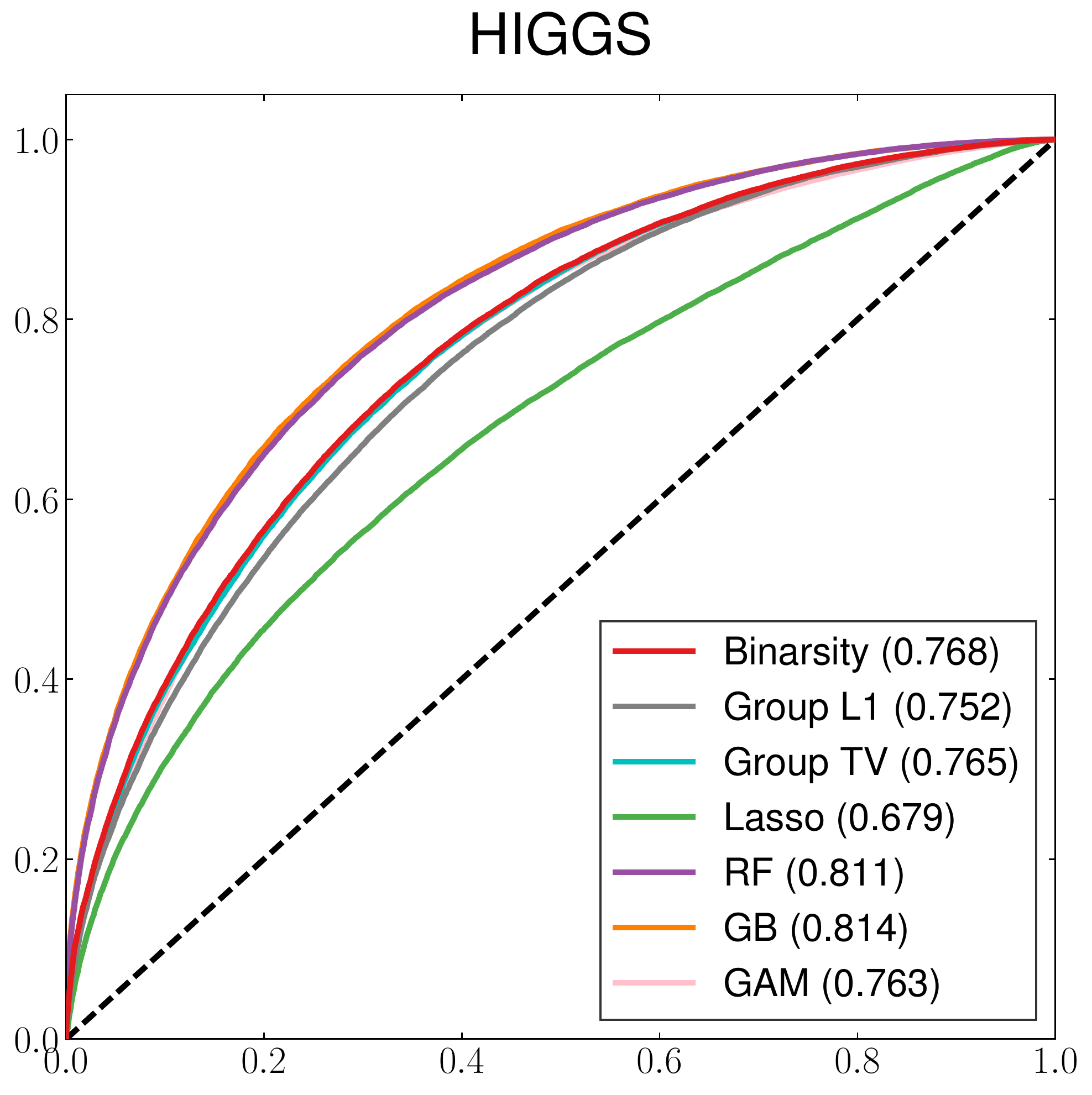}

\caption{\small Performance comparison using ROC curves and AUC scores (given between parenthesis) computed on test sets. The 4 last datasets contain too many examples for SVM (RBF kernel). Binarsity consistently does a better job than Lasso, Group L1, Group TV and GAM. Its performance is comparable to SVM, RF and GB but with computational timings that are orders of magnitude faster, see Figure~\ref{fig:computing-time-comparison}.}
\label{fig:roc-curves}
\end{figure*}

\begin{figure*}
\centering
\includegraphics[width=\textwidth]{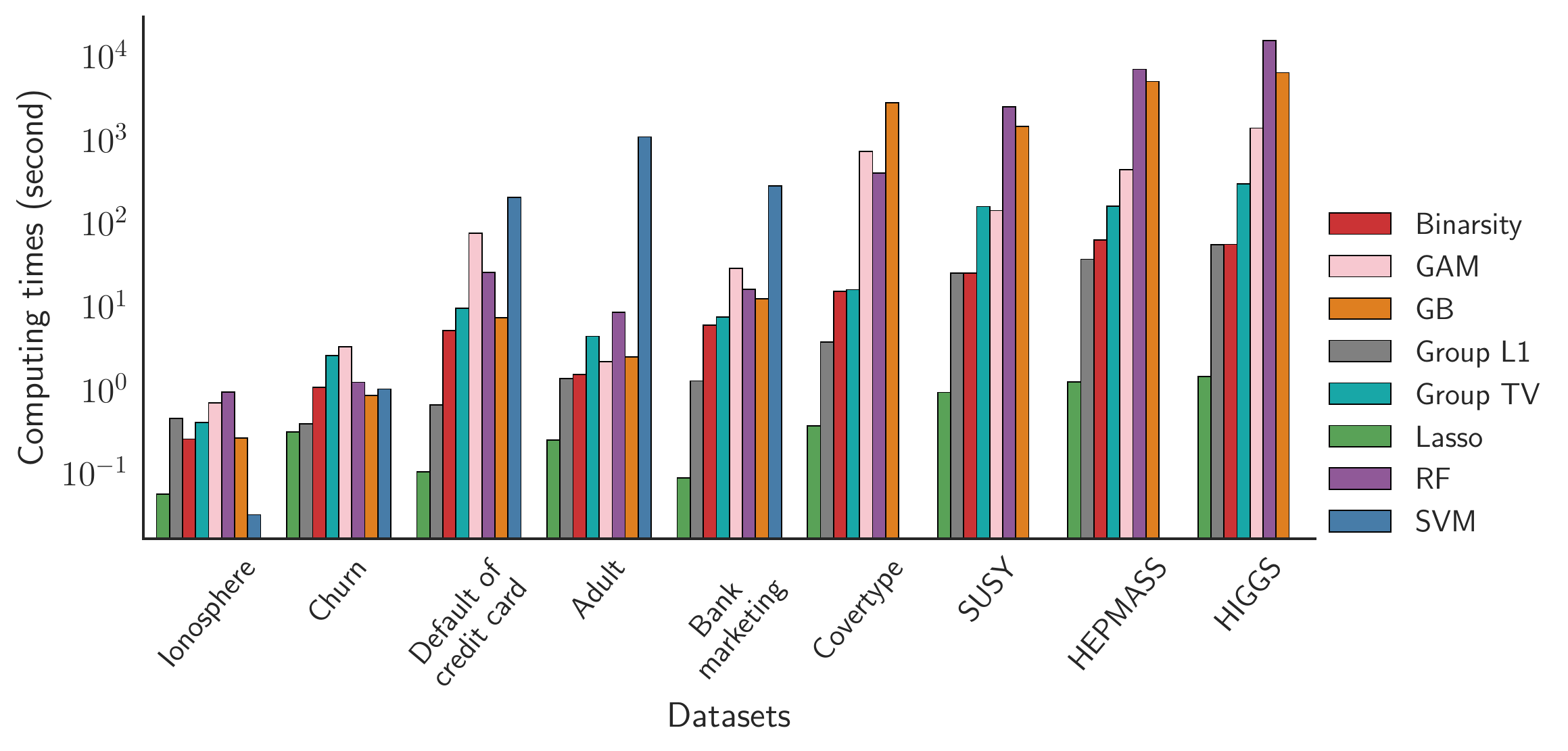}
\caption{\small Computing time comparisons (in seconds) between the methods on the considered datasets. Note that the time values are $\log$-scaled. These timings concern the learning task for each model with the best hyper parameters selected, after the cross validation procedure. The 4 last datasets contain too many examples for the SVM with RBF kernel to be trained in a reasonable time. Roughly, binarsity is between 2 and 5 times slower than $\ell_1$ penalization on the considered datasets, but is more than 100 times faster than random forests or gradient boosting algorithms on large datasets, such as HIGGS.}
\label{fig:computing-time-comparison}
\end{figure*}

The results of all our experiments are reported in Figures~\ref{fig:roc-curves} and~\ref{fig:computing-time-comparison}.
In Figure~\ref{fig:roc-curves} we compare the performance of binarsity with the baselines on all 9 datasets, using ROC curves and the Area Under the Curve (AUC), while we report computing (training) timings in Figure~\ref{fig:computing-time-comparison}.
We observe that binarsity consistently outperforms Lasso, as well as Group L1: this highlights the importance of the TV norm within each group. 
The AUC of Group TV is always slightly below the one of binarsity, and more importantly it involves a much larger training time: convergence is slower for Group TV, since it does not use the linear constraint of binarsity, leading to a ill-conditioned problem (sum of binary features equals 1 in each block).
Finally, binarsity outperforms also GAM and its performance is comparable in all considered examples to RF and GB, with computational timings that are orders of magnitude faster, see Figure~\ref{fig:computing-time-comparison}.
All these experiments illustrate that binarsity achieves an extremely competitive compromise between computational time and performance, compared to all considered baselines.

\section{Conclusion}
\label{section:discussion}

In this paper, we introduced the binarsity penalization for one-hot encodings of continuous features. 
We illustrated the good statistical properties of binarsity for generalized linear models by proving non-asymptotic oracle inequalities.
We conducted extensive comparisons of binarsity with state-of-the-art algorithms for binary classification on several standard datasets.
Experimental results illustrate that binarsity significantly outperforms Lasso, Group L1 and Group TV penalizations and also generalized additive models, while being competitive with random forests and boosting. 
Moreover, it can be trained orders of magnitude faster than boosting and other ensemble methods. 
Even more importantly, it provides interpretability.
Indeed, in addition to the raw feature selection ability of binarsity, the method pinpoints significant cut-points for all continuous feature.
This leads to a much more precise and deeper understanding of the model than the one provided by Lasso on raw features.
These results illustrate the fact that binarsity achieves an extremely competitive compromise between computational time and performance, compared to all considered baselines.

\section{Proofs}

In this Section we gather the proofs of all the theoretical results proposed in the paper.
Throughout this Section, we denote by $\partial(\phi)$ the subdifferential mapping of a convex function $\phi.$

\subsection{Proof of Proposition~\ref{proposition:prox-btv-primal}}
\label{appendix:proof-of-proposition:prox-btv-primal}

Recall that the indicator function $\delta_j$ is given by~\eqref{eq:def_delta_j}.
For any fixed $j=1, \ldots, p,$ we prove that $\prox_{\norm{\cdot}_{\TV,\hat w_{j,\bullet}} + \delta_j}$ is the composition of $\prox_{\norm{\cdot}_{\TV,\hat w_{j,\bullet}}}$ and $\prox_{\delta_j},$ namely
\begin{equation*}
\prox_{\norm{\cdot}_{\TV,\hat w_{j,\bullet}} + \delta_j}(\theta_{j,\bullet}) = \prox_{\delta_j}\big(\prox_{\norm{\cdot}_{\TV,\hat w_{j,\bullet}}}(\theta_{j,\bullet})\big)
\end{equation*}
for all $\theta_{j,\bullet} \in \R^{d_j}$. 
Using Theorem~1 in~\cite{Yu-13}, it is sufficient to show that for all $\theta_{j,\bullet} \in \R^{d_j},$ we have  
\begin{equation}
\label{sub-diff-inclusion}
\partial \big(\norm{\theta_{j,\bullet}}_{\TV,\hat w_{j,\bullet}}\big) \subseteq \partial \big(\norm{\prox_{\delta_j}(\theta_{j,\bullet})}_{\TV, \hat w_{j,\bullet}}\big).
\end{equation} 
We have $\prox_{\delta_j} (\theta_{j,\bullet}) = \Pi_{\text{span}\{n_j\}^\perp}(\theta_{j,\bullet}),$ where $\Pi_{\text{span}\{n_j\}^\perp}(\cdot)$ stands for the projection onto the orthogonal of $\text{span}\{n_j\}$.
This projection simply writes
\begin{equation*}
  \Pi_{\text{span}\{n_j\}^\perp}(\theta_{j,\bullet}) = \theta_{j, \bul} - 
  \frac{n_j^\top \theta_{j, \bullet}}{\norm{n_j}_2^2} n_j
\end{equation*}
Now, let us define the $d_j \times d_j$ matrix $D_j$ by
\begin{equation}
\label{eq:D_j-matrix-definition}
 {{D}_{j}}=
\begin{bmatrix}
1  & 0 & & 0 \\
-1 & 1 &       \\
&  \ddots& \ddots \\
0 & & -1& 1 
\end{bmatrix} \in \R^{d_j}\times \R^{d_j}.
\end{equation}
We then remark that for all $\theta_{j,\bullet} \in \R^{d_j}$,
\begin{equation}
\label{eq:weighted-TV-hadamard-expr}
\norm{\theta_{j,\bullet}}_{\TV, \hat w_{j,\bullet}} 
= \sum_{k=2}^{d_j} \hat w_{j,k} |\theta_{j,k} - \theta_{j, k-1}|  
= \norm{\hat w_{j,\bullet} \odot D_j \theta_{j,\bullet}}_{1}.
\end{equation}
Using subdifferential calculus (see details in the proof of Proposition~\ref{proposition: KKT-conditions} below), one has 
\begin{equation*}
\partial \big(\norm{\theta_{j,\bullet}}_{\TV, \hat w_{j,\bullet}}\big) = \partial \big(\norm{\hat w_{j,\bullet}\odot{{D_{j}}}\theta_{j,\bullet}}_{1}\big) = {{D_{j}}}^\top \hat w_{j,\bullet}\odot \sgn({{D_j}}\theta_{j,\bullet}).
\end{equation*}
Then, the linear constraint $n_j^\top \theta_{j, \bul} = 0$ entails 
\begin{equation*}
{{D_j}}^\top \hat w_{j,\bullet}\odot \sgn({{D_j}} \theta_{j,\bullet}) = {{D_j}}^\top 
\hat w_{j,\bullet}\odot \sgn \Big( {{D_j}} \big( \theta_{j,\bullet} - 
\frac{n_j^\top \theta_{j, \bullet}}{\norm{n_j}_2^2} n_j \big) \Big),
\end{equation*}
which leads to~\eqref{sub-diff-inclusion} and concludes the proof of the Proposition. $\hfill \square$

\subsection{Proximal operator of the weighted TV penalization}
\label{appendix:proximal-operator-wTV}

We recall in Algorithm~\ref{algorithm-weighted-TV-agg} an algorithm provided in~\cite{alaya2014} for the computation of the proximal operator of the weighted total-variation penalization
\begin{equation}
\label{primal-prox-wTV}
\beta =\prox_{\norm{\cdot}_{\TV,\hat w}}(\theta) \in \argmin_{\theta\in \R^m} \Big\{ \frac{1}{2} \norm{\beta
    - \theta}_2^2 + \norm{\theta}_{\TV, \hat w} \Big\}.
\end{equation}
A quick explanation of this algorithm is as follows. 
The algorithm runs forwardly through the input vector $(\theta_1, \ldots, \theta_m).$ 
Using Karush-Kuhn-Tucker (KKT) optimality conditions~\cite{BoyVan-04}, we have that at a location $k,$ 
the weight $\beta_k$ stays constant whenever $|u_k| < \hat w_{k+1}$,
where $u_k$ is a solution to a dual problem associated to the primal problem~\eqref{primal-prox-wTV}.
If not possible, it goes back to the last location where a jump can be introduced in $\beta$, validates the current segment until this location, starts a new segment, and continues.
\LinesNotNumbered
\begin{algorithm}[htbp]
\SetNlSty{textbf}{}{.}
\DontPrintSemicolon
 \caption{\small Proximal operator of weighted TV penalization}
\label{algorithm-weighted-TV-agg}
 \KwIn{vector $\theta=\big(\theta_1, \ldots, \theta_m\big) ^\top\in \R^m$ and weights $\hat w = (\hat w_1,\ldots, \hat w_m) \in \R^{m}_{+}.$}
 \KwOut{vector $\beta = \prox_{\norm{\cdot}_{\TV,\hat w}}(\theta)$} 
 \nl\textbf{Set} {$k=k_0=k_-=k_+ \gets 1$\\$\qquad \beta_{\min}  \gets \theta_1- \hat w_2\ ;\ \beta_{\max} \gets \theta_1+ \hat w_2$\\$\qquad u_{\min}\gets \hat w_2\ ;\ u_{\max} \gets -\hat w_2$}\\
 \nl \label{step2}{\If{$k=m$}
{  $\beta_m \gets \beta_{\min}+ u_{\min}$}}
\nl \label{step3} \If (\tcc*[f]{negative jump}){$\theta_{k+1} + u_{\min} < \beta_{\min} - \hat w_{k+2}$}
{$\beta_{k_0}= \cdots =\beta_{k_-} \gets \beta_{\min}$\\
 $k=k_0=k_-=k_+\gets k_- + 1$\\
 $ \beta_{\min} \gets \theta_{k} - \hat w_{k +1}+ \hat w_{k }\ ;\ \beta_{\max}   \gets \theta_{k} + \hat w_{k +1}+ \hat w_{k}$\\
 $u_{\min}   \gets \hat w_{k +1}\ ;\ u_{\max} \gets -\hat w_{k+1}$
}
 \nl \ElseIf (\tcc*[f]{positive jump}){$\theta_{k+1} + u_{\max} > \beta_{\max}+ \hat w_{k+2}$}
{$\beta_{k_0}= \ldots =\beta_{k_+}  \gets \beta_{\max}$\\
$k=k_0=k_-=k_+\gets k_+ + 1$\\
$\beta_{\min} \gets \theta_{k} - \hat w_{k+1} - \hat w_{k }\ ;\ \beta_{\max}\gets \theta_{k} +\hat w_{k+1}-  \hat w_{k}$\\
$ u_{\min}   \gets \hat w_{k+1}\ ;\ u_{\max} \gets -\hat w_{k+1}$
}
 \nl \Else (\tcc*[f]{no jump}){ \textbf{set }$k \gets k+1$\\
 $\qquad u_{\min} \gets  \theta_{k} + \hat w_{k+1} - \beta_{\min}$\\
 $\qquad u_{\max} \gets  \theta_{k} - \hat w_{k+1} - \beta_{\max}$
  \If{$u_{\min}  \geq \hat w_{k+1}$}{ $\beta_{\min}\gets \beta_{\min}+ \frac{u_{\min} - \hat w_{k+1}}{k-k_0+1}$\\
$u_{\min} \gets \hat w_{k+1}$\\
$k_- \gets k$}
  \If{$u_{\max} \leq -\hat w_{k+1}$}{$\beta_{\max}\gets \beta_{\max}+ \frac{u_{\max} + \hat w_{k+1}}{k-k_0+1}$\\
$u_{\max} \gets  -\hat w_{k+1}$\\
$k_+ \gets k$}
}
\nl \If{$ k< m$ } { go to \textbf{\ref{step3}.}}
\nl \If{$u_{\min} < 0$ } {$\beta_{k_0}= \cdots =\beta_{k_-} \gets \beta_{\min}$\\
$k=k_0=k_- \gets k_-+1$\\
$\beta_{\min}\gets \theta_k - \hat w_{k+1}+ \hat w_{k}$\\
$u_{\min}\gets \hat w_{k+1}\ ;\ u_{\max} \gets \theta_k+ \hat w_{k} - v_{\max}$
\\go to \textbf{\ref{step2}.}}
 \nl \ElseIf{$u_{\max} > 0$ } { $ \beta_{k_0}= \cdots =\beta_{k_+}\gets \beta_{\max}$\\
$k=k_0=k_+ \gets k_++1$\\
$\beta_{\max} \gets \theta_k +\hat w_{k+1} -  \hat w_{k}$\\
$u_{\max}\gets - \hat w_{k+1}\ ;\ u_{\min}\gets \theta_k- \hat w_{k} - u_{\min}$ \\ 
go to \textbf{\ref{step2}.}}
\nl \Else{$\beta_{k_0}= \cdots =\beta_m\gets \beta_{\min} + \frac{u_{\min}}{k-k_0+1}$}
\end{algorithm}

\subsection{Proof of Theorem~\ref{thm:oracle}}
\label{proof-fast-oracle-ineq-bina}

The proof relies on several technical properties that are described below.
From now on, we consider $\by = [y_1 \cdots y_n]^\top$, $\bX = [x_1 \cdots x_n]^\top$, $m^0(\bX) = [m^0(x_1) \cdots m^0(x_n)]^\top,$ and recalling that $m_\theta(x_i) = \theta^\top x^B_i)$ we introduce $m_\theta(\bX) = [m_\theta(x_1) \cdots m_\theta(x_n) ]^\top$ and $b'(m_\theta(\bX)) = [b'(m_\theta(x_1)) \cdots b'(m_\theta(x_n)) ]^\top.$

Let us now define the Kullback-Leibler divergence between the true probability density funtion $f^0$ defined in~\eqref{distribut-glm} and a candidate $f_\theta$ within the generalized linear model $f_\theta(y|x) = \exp\big(ym_\theta(x) - b(m_\theta(x))$ as follows
\begin{align*}
\mathrm{KL}_n(f^0, f_\theta) &= \frac{1}{n} \sum_{i=1}^n \E_{\Pyx}\Big[\log\frac{f^0(y_i|x_i)}{f_{\theta}(y_i | x_i)}\Big]\\
&:= \mathrm{KL}_n(m^0(\bX), m_\theta (\bX)),
\end{align*}
where $\Pyx$ is the joint distribution of $\by$ given $\bX$.
We then have the following Lemma.
\begin{lemma}
\label{lemma-excess-risk-KL}
The excess risk satisfies
\begin{equation*}
  R(m_{\theta}) - R(m^0) = \phi \mathrm{KL}_n(m^0(\bX), m_\theta (\bX)),
\end{equation*}
where we recall that $\phi$ is the dispertion parameter of the generalized linear model, see~\eqref{distribut-glm}. 
\end{lemma}

\medskip
\noindent
\textbf{Proof.}
If follows from the following simple computation
\begin{align*}
&\mathrm{KL}_n(m^0(\bX), m_\theta (\bX))\\
&\qquad= \phi^{-1} \frac 1n \sum_{i=1}^n \E_{\Pyx} \Big[\big( -y_i m_{\theta}(x_i) +b(m_{\theta}(x_i))\big) - \big(-y_i m^0(x_i) +b(m^0(x_i))\big)  \Big]\\
&\qquad = \phi^{-1} \big(R(m_{\theta}) - R(m^0)\big)
\end{align*}
which proves the Lemma. $\hfill \square$

\subsection{Optimality conditions}

As explained in the following Proposition, a solution to problem~\eqref{model:general} can be characterized using the Karush-Kuhn-Tucker (KKT) optimality conditions~\cite{BoyVan-04}.
\begin{proposition}
\label{proposition: KKT-conditions}
A vector $\hat \theta = [{\hat\theta}_{1,\bullet}^\top \cdots {\hat\theta}_{p,\bullet}^\top]^\top \in \R^d$ is 
an optimum of the objective function~\eqref{model:general} if and only if there are subgradients 
$\hat h = [\hat h _{j, \bullet}]_{j=1, \ldots, p} \in \partial \norm{\hat \theta}_{ \TV, \hat w}$ and 
$\hat g  = [\hat g _{j, \bullet}]_{j=1, \ldots, p} \in 
\partial [\delta_j(\hat \theta_{j,\bullet})]_{j=1, \ldots, p}$ such that 
\begin{equation*}
 \nabla R_n(\hat \theta_{j,\bullet})+{\hat h}_{j,\bullet}  + {\hat g}_{j,\bullet} = \mathbf{0}, 
\end{equation*}
where
\begin{equation}
\label{subdifferential-of-TVw}
\left\{
   \begin{array}{ll}
       {\hat h}_{j,\bullet} = D_{{j}}^\top \big(\hat w_{j,\bullet}\odot\sgn(D_{{j}}{\hat\theta}_{j,\bullet})\big) & \mbox{if }  j \in  J(\hat \theta),\\
       {\hat h}_{j,\bullet} \in  D_{{j}}^\top \big( \hat w_{j,\bullet}\odot {[-1,+1]}^{d_j}\big) & \mbox{if }  j \in   J^\complement(\hat \theta),
 \end{array} 
\right.
\end{equation}
and where we recall that $J(\hat \theta)$ is the support set of $\hat \theta$. 
The subgradient $\hat g_{j, \bullet}$ belongs to 
\begin{equation*}
\partial\big(\delta_j(\hat \theta_{j,\bullet})\big) = \big\{\mu_{j, \bullet}\in \R^{d_j}: \mu_{j, \bullet}^\top 
\theta_{j, \bullet} \leq \mu_{j, \bullet}^\top \hat \theta_{j,\bullet} \; \text{ for all } \; 
\theta_{j, \bullet} \; \text{ such that } \; n_j^\top \theta_{j,\bullet} = 0 \big\}.
\end{equation*}
For the generalized linear model, we have 
\begin{equation}
\label{KKT-logistic}
\normalfont
\frac 1n \big( \XB_{\bullet,j}\big)^\top\big(b'(m_{\hat\theta}(\bX))- \by\big) + {\hat h}_{j,\bullet} + \hat g_{j,\bullet} + \hat f_{j,\bullet}= \mathbf{0},
\end{equation}
where $\hat f = [{\hat f }_{j,\bullet}]_{j=1, \ldots, p}$ belongs to the normal cone of the ball $B_d(\rho).$
\end{proposition}

\medskip
\noindent
\textbf{Proof.}
The function $\theta \mapsto R_n(\theta)$ is differentiable, so the subdifferential of $R_n(\cdot) +  \bina(\cdot)$ at a point $\theta = (\theta_{j,\bullet})_{j=1, \ldots, p} \in \R^d$ is given by
\begin{equation*}
\partial (R_n(\theta) + \bina(\theta)) = \nabla R_n(\theta) + \partial(\bina(\theta)),
\end{equation*}
where 
$\nabla R_n(\theta) = \Big[\frac{\partial R_n(\theta)}{\partial \theta_{1,\bullet}} \cdots 
\frac{\partial R_n(\theta)}{\partial \theta_{p,\bullet}}\Big]^\top$
and
\begin{equation*}
\partial \bina(\theta) =
\Big[ \partial \norm{\theta_{1,\bullet}}_{ \TV, \hat w_{1,\bullet}} + \partial \delta_j(\theta_{1,\bullet})  
\; \cdots \; \partial \norm{\theta_{p,\bullet}}_{ \TV, \hat w_{p,\bullet}} + \partial \delta_j(\theta_{p,\bullet})  \Big]^\top.
\end{equation*}
We have $\norm{\theta_{j,\bullet}}_{ \TV, \hat w_{j,\bullet}} = \norm{ \hat w_{j,\bullet} \odot D_j\theta_{j,\bullet}}_1$ for all $j =1, \ldots,p$. Then, by applying some properties of the subdifferential calculus, we get
\begin{equation}
\label{subdifferential-btv}
\partial \norm{\theta_{j,\bullet}}_{ \TV, \hat w_{j,\bullet}}  
= \begin{cases}
D_j^\top \sgn(\hat w_{j,\bullet} \odot D_j\theta_{j,\bullet}) & \text{ if } D_j\theta \neq \mathbf{0}, \\
D_j^\top\big(\hat w_{j,\bullet}\odot v_j) & \text{ otherwise},
\end{cases}
\end{equation}
where $v_j \in [-1,+1]^{d_j}$ for all $j=1, \ldots, p$.
For generalized linear models, we rewrite 
\begin{equation}
\label{glm-model-in-proof}
\hat \theta \in \argmin_{\theta \in \R^d } \big\{R_n(\theta) + \bina(\theta) + \delta_{B_d(\rho)}(\theta)\big\},
\end{equation}
where $\delta_{B_d(\rho)}$ is the indicator function of $B_d(\rho)$.
Now, $\hat \theta = [{\hat\theta}_{1,\bullet}^\top \cdots {\hat\theta}_{p,\bullet}^\top]^\top$ is an optimum of~\eqref{glm-model-in-proof} if and only if $\mathbf{0} \in \nabla R_n(m_{\hat \theta}) 
+ \partial \norm{\hat \theta}_{\TV, \hat w} + \partial \delta_{B_d(\rho)}(\hat \theta_{})$.
Recall that the subdifferential of $\delta_{B_d(\rho)}(\cdot)$ is the normal cone of $B_d(\rho)$, namely
\begin{equation}
\label{normal-cone}
\partial \delta_{B_d(\rho)}(\hat \theta) = \big\{\eta \in \R^d : 
\eta^\top \theta \leq \eta^\top \hat \theta \text{ for all } \theta \in B_d(\rho) \}.
\end{equation}
One has
\begin{equation}
\label{nabla-logistic}
\frac{\partial R_n(\theta)}{\partial \theta_{j,\bullet}} 
 = \frac{1}{n} (\XB_{\bullet,j})^\top(b'(m_{\hat\theta}(\bX)) - \by),
\end{equation}
so that together with~\eqref{nabla-logistic} and~\eqref{normal-cone} we obtain~\eqref{KKT-logistic}, which concludes the proof of Proposition~\ref{proposition: KKT-conditions}. $\hfill \square$

\subsection{Compatibility conditions}
\label{sub:compatibility_conditions}

Let us define the block diagonal matrix ${{\bf{D}}} = \diag({D}_{{1}}, \ldots, {D}_{{p}})$ with $D_j$
 defined in~\eqref{eq:D_j-matrix-definition}. 
We denote its inverse ${{T}_{j}}$ which is defined by the $d_j \times d_j$ lower triangular matrix 
with entries $({{T}_{j}})_{r,s} = 0$ if $r < s$ and $({{T}_{j}})_{r,s} = 1$ otherwise.
We set $ {{\bf{T}}} =\diag({T}_{{1}}, \ldots, {T}_{{p}})$, so that one has ${\bf{D}} ^{-1} = {\bf{T}}$.

In order to prove Theorem~\ref{thm:oracle}, we need the following results which give a compatibility property~\cite{vandegeer2008,vandegeer2013,DalHeiLeb14} for the matrix ${\bf{T}}$, see
 Lemma~\ref{lemma-compatibility-Tw} below and for the matrix $\XB{\bf{T}}$, see Lemma~\ref{lemma:CC-XBTw} below.
For any concatenation of subsets $ K=[K_1, \ldots, K_p],$ we set
\begin{equation}
\label{defn:concatenation}
 K_j = \{\tau_j^1, \ldots, \tau_j^{b_j}\} \subset \{1, \ldots, d_j\}
\end{equation}
for all $j=1, \ldots, p$ with the convention that $\tau_j^0 = 0$ and $\tau_j^{{b_j} +1} = d_j +1$.
\begin{lemma}
\label{lemma-compatibility-Tw}
Let $ \gamma\in \R^d_+$ be given and $K = [K_1, \ldots, K_p]$ with $K_j$ given by~\eqref{defn:concatenation} for all $j=1, \ldots, p$.
Then, for every $u \in \R^d \backslash \{\mathbf{0}\}$, we have 
\begin{equation*}
\frac{\norm{{\bf{T}} u}_2}{|\norm{u_K\odot \gamma_K}_1 - \norm{u_{K^\complement} \odot \gamma_{K^\complement}}_1|}  \geq  \kappa_{{\bf{T}},\gamma}(K),
\end{equation*}
where 
\begin{equation*}
 \kappa_{{\bf{T}},\gamma}(K) = \bigg\{ 32  \sum_{j=1}^p\sum_{k=1}^{d_j} | \gamma_{j,k+1} -\gamma_{j,k}|^2+  2|K_j|\norm{\gamma_{j,\bullet}}_\infty^2\Delta_{\min, K_j}^{-1}\bigg\}^{-1/2},
\end{equation*}
and $\Delta_{\min, K_j} = \min_{r=1, \ldots b^j}| \tau_j^{{r_j}} - \tau_j^{{r_j} -1 }|.$
\end{lemma}

\medskip
\noindent
\textbf{Proof.}
Using Proposition~3 in~\cite{DalHeiLeb14}, we have
\begin{equation*}
\begin{split}
&\norm{u_K\odot \gamma_K}_1 - \norm{u_{K^\complement} \odot \gamma_{K^\complement}}_1\\
&\qquad \qquad = \sum_{j=1}^p\norm{u_{K_j}\odot \gamma_{K_j}}_1 - \norm{u_{{K_j}^\complement} \odot \gamma_{{K_j}^\complement}}_1 \\
&\qquad \qquad \leq \sum_{j=1}^p 4\norm{T_{{j}} u_{j,\bullet}}_2 \bigg\{2\sum_{k=1}^{d_j} |  \gamma_{j,k+1} -\gamma_{j,k}|^2
+ 2(b_j +1) \norm{\gamma_{j,\bullet}}_\infty^2\Delta_{\min, K_j}^{-1}\bigg\}^{1/2}.
\end{split}
\end{equation*}
Using H\"older's inequality for the right hand side of the last inequality gives 
\begin{equation*}
\begin{split}
&\norm{u_K\odot \gamma_K}_1 - \norm{u_{K^\complement} \odot \gamma_{K^\complement}}_1\\
&\qquad \qquad \leq \norm{{\bf{T}} u}_2 \bigg\{ 32\sum_{j=1}^p \sum_{k=1}^{d_j} |  \gamma_{j,k+1} -\gamma_{j,k} |^2 + 2|K_j| \norm{\gamma_{j,\bullet}}_\infty^2\Delta_{\min, K_j}^{-1}\bigg\}^{1/2},
\end{split}
\end{equation*}
which completes the proof of the Lemma. $\hfill \square$

\medskip

Combining Assumption~\ref{assumption:RE-XB} and Lemma~\ref{lemma-compatibility-Tw} allows to establish a compatibility condition satisfied by $\XB {\bf{T}}$.
\begin{lemma} 
\label{lemma:CC-XBTw}
Let $\gamma \in \R^d_+$ be given and $K = [K_1, \ldots, K_p]$ with $K_j$ given by~\eqref{defn:concatenation} for $j=1, \ldots, p$.
Then, if Assumption~\ref{assumption:RE-XB} holds, one has   
\begin{equation}
\label{ineq:compatibility-XT}
\inf\limits_{\substack{u \in \mathscr{C}_{1,\hat w}(K)\backslash\{\mathbf{0}\}}}\Big\{\frac{\norm{\XB {\bf{T}} u}_2}{\sqrt{n} \;|\; \norm{u_K\odot \gamma_K}_1 - \norm{u_{K^\complement} \odot \gamma_{K^\complement}}_1 |} \Big\}\geq  \kappa_{{\bf{T}},\gamma}(K)\kappa(K),
\end{equation}
where 
\begin{equation}
\label{C-1}
\mathscr{C}_{1, \hat w}(K) \stackrel{}{=} \Big\{u \in \R^d: \sum_{j=1}^p \norm{(u_{j, \bullet})_{{K_j}^\complement}}_{1,\hat w_{j,\bullet}}  \leq 2\sum_{j=1}^p \norm{(u_{j, \bullet})_{K_j}}_{1,\hat w_{j,\bullet}}\Big\}.
\end{equation}
\end{lemma}

\noindent
\textbf{Proof.}
Lemma~\ref{lemma-compatibility-Tw} gives
\begin{equation*}
\frac{\norm{\XB {\bf{T}}u}_2}{\sqrt{n} |\norm{u_K\odot \gamma_K}_1 - \norm{u_{K^\complement} \odot \gamma_{K^\complement}}_1|} \geq \kappa_{{\bf{T}},\gamma}(K)\frac{\norm{\XB {\bf{T}}u}_2}{\sqrt{n} \norm{{\bf{T}}u}_2}.
\end{equation*}
Now, we note that if $u \in \mathscr{C}_{1, \hat w}(K)$, then ${\bf{T}} u \in  \mathscr{C}_{ \TV, \hat w}(K).$ Hence, Assumption~\ref{assumption:RE-XB} entails
\begin{equation*}
\frac{\norm{\XB {\bf{T}} u}_2}{\sqrt{n} |\norm{u_K\odot \gamma_K}_1 - \norm{u_{K^\complement} \odot \gamma_{K^\complement}}_1|} \geq \kappa_{{\bf{T}},\gamma}(K) \kappa(K),
\end{equation*}
which concludes the proof of the Lemma. $\hfill \square$

\subsection{Connection between the empirical Kullback-Leibler divergence and the empirical squared norm}
\label{subsection-connection-betwen-KL-sqnorm}

The next Lemma is from~\cite{bach2010selfconcordance} (see Lemma~1 herein).
\begin{lemma}
\label{lemma:self-concordance}
Let $\varphi:\R \rightarrow \R$ be a three times differentiable convex function such that for all $t\in\R,$ $|\varphi'''(t)| \leq M |\varphi''(t)|$ for some $M \geq 0.$ Then, for all $t \geq 0$, one has
\begin{equation*}
\frac{\varphi''(0)}{M^2}\psi(-Mt) \leq \varphi(t) - \varphi(0) - \varphi'(0)t \leq \frac{\varphi''(0)}{M^2}\psi(Mt),
\end{equation*}
with $\psi(u) = e^u - u - 1$.
\end{lemma}

This Lemma entails the following in our setting.
\begin{lemma}
\label{lemma-connection-L2-KL} 
Under Assumption~\ref{ass:glm}, one has
\begin{align*}
\frac{L_n \psi(-2(C_n + \rho))}{4 \phi (C_n + \rho)^2} \frac{1}{n}\norm{m^0(\bX) - m_\theta(\bX)}_2^2 
&\leq \mathrm{KL}_n(m^0(\bX), m_\theta(\bX)), \\
\frac{U_n \psi(2(C_n + \rho))}{4 \phi (C_n + \rho)^2} \frac{1}{n}\norm{m^0(\bX) - m_\theta(\bX)}_2^2 
&\geq \mathrm{KL}_n(m^0(\bX), m_\theta(\bX)),
\end{align*}
for all $\theta \in B_d(\rho)$.
\end{lemma}

\noindent
\textbf{Proof.}
Let us consider the function $G_n:\R \rightarrow \R$ defined by $G_n(t)= R_n(m^0 + t m_\eta)$, with $m_\eta$ to be defined later, which writes
\begin{equation*}
G_n(t) = \frac 1n \sum_{i=1}^n b(m^0(x_i) + tm_\eta(x_i)) - \frac 1n\sum_{i=1}^n y_i (m^0(x_i) + tm_\eta(x_i)).
\end{equation*}
We have
\begin{equation*}
\begin{split}
&G'_n(t)= \frac{1}{n} \sum_{i=1}^n m_\eta(x_i) b'(m^0(x_i) + tm_\eta(x_i)) - \frac{1}{n} \sum_{i=1}^n y_i m_\eta(x_i),\\
&G''_n(t) = \frac{1}{n} \sum_{i=1}^n m^2_\eta(x_i) b''(m^0(x_i) + tm_\eta(x_i)),\\
\text{and} \quad &G'''_n(t) = \frac{1}{n} \sum_{i=1}^n m^3_\eta(x_i) b'''(m^0(x_i) + tm_\eta(x_i)).
\end{split}
\end{equation*}
Using Assumption~\ref{ass:glm}, we have $|G'''_n(t)| \leq C_b \norm{m_\eta}_{\infty}|G''_n(t)|$ where $\norm{m_\eta}_{\infty} := \max\limits_{i=1, \ldots, n}|m_\eta(x_i)|$.
Lemma~\ref{lemma:self-concordance} with $M = C_b \norm{m_\eta}_{\infty}$ gives
\begin{equation*}
G''_n(0)\frac{ \psi(- C_b \norm{m_\eta}_{\infty}t)}{C_b^2\norm{m_\eta}_{\infty}^2}\leq G_n(t) - G_n(0) 
- tG'_n(0) \leq G''_n(0) \frac{ \psi(C_b\norm{m_\eta}_{\infty}t)}{C_b^2 \norm{m_\eta}_{\infty}^2}
\end{equation*}
for all $t\geq 0$ and $t=1$ leads to
\begin{equation*}
G''_n(0)\frac{ \psi(-C_b \norm{m_\eta}_{\infty})}{C_b^2 \norm{m_\eta}_{\infty}^2}
\leq R_n(m^0 +  m_\eta) - R_n(m^0) - G'_n(0) \leq 
G''_n(0) \frac{ \psi(C_b \norm{m_\eta}_{\infty})}{C_b^2 \norm{m_\eta}_{\infty}^2}.%
\end{equation*}
An easy computation gives
\begin{align*}
- G'_n(0) = \frac{1}{n} \sum_{i=1}^n m_\eta(x_i) \big(y_i - b'(m^0(x_i)) \big)
\; \text{ and } \; G''_n(0) = \frac{1}{n} \sum_{i=1}^n m^2_\eta(x_i)b''(m_{\eta }(x_i)),
\end{align*}
and since obviously $\E_{\Pyx}[G_n'(0)] = 0$, we obtain
\begin{align*}
&G''_n(0)\frac{ \psi(-C_b \norm{m_\eta}_{\infty})}{C_b^2 \norm{m_\eta}_{\infty}^2}
\leq R(m^0 + m_\eta) - R(m^0)
\leq G''_n(0)\frac{ \psi(C_b \norm{m_\eta}_{\infty})}{C_b^2 \norm{m_\eta}_{\infty}^2}.
\end{align*}
Now, choosing $m_{\eta} = m_\theta - m^0$ and combining Assumption~\ref{ass:glm} 
with Equation~\eqref{lemma:control_inner_ball} gives
\begin{equation*}
C_b \norm{m_\eta}_{\infty} \leq C_b  \max_{i=1, \ldots, n} (|\inr{x_i^B, \theta}| + |m^0(x_i)| ) \leq C_b (\rho  + C_n).
\end{equation*}
Hence, since $x \mapsto \psi(x) / x^2$ is an increasing function on $\R^+$, we end up with
\begin{align*}
G''_n(0)\frac{ \psi(-C_b (C_n + \rho))}{C_b^2 (C_n + \rho)^2}
&\leq R(m_{\theta}) - R(m_{0}) = \phi \mathrm{KL}_n(m^0(\bX), m_{\theta}(\bX)),\\
G''_n(0)\frac{ \psi(C_b (C_n + \rho))}{C_b^2 (C_n + \rho)^2}
&\geq R(m_{\theta}) - R(m_{0})= \phi \mathrm{KL}_n(m^0(\bX), m_{\theta}(\bX)),
\end{align*}
and since $G''_n(0) = n^{-1} \sum_{i=1}^n (m_\theta(x_i) - m^0(x_i) )^2 b''(m^0(x_i))$, 
we obtain
\begin{align*}
\frac{L_n \psi(-C_b (C_n + \rho))}{C_n^2 \phi (C_n + \rho)^2} \frac{1}{n} \norm{m^0(\bX) - m_\theta(\bX)}_2^2
&\leq \mathrm{KL}_n(m^0(\bX), m_{\theta}(\bX)), \\
\frac{U_n\psi(C_b (C_n + \rho))}{C_b^2 \phi (C_n + \rho)^2} \frac{1}{n} \norm{m^0(\bX) - m_\theta(\bX)}_2^2 
&\geq \mathrm{KL}_n(m^0(\bX), m_{\theta}(\bX)),
\end{align*}
which concludes the proof of the Lemma. $\hfill \square$

\subsection{Proof of Theorem~\ref{thm:oracle}}
\label{appendix:proof-theorem-oracle-logistic-V2}

Let us recall that
\begin{equation*}
R_n(m_\theta) = \frac{1}{n} \sum_{i=1}^n  b(m_\theta(x_i))- \frac{1}{n} \sum_{i=1}^n y_i m_{\theta}(x_i)
\end{equation*}
for all $\theta \in \R^d$ and that
\begin{equation}
\label{proof-def-estimator-logistic}
\hat \theta \in \argmin_{\theta \in B_d(\rho)} \big\{R_n(\theta) + \bina(\theta)\big\}.
\end{equation}
Proposition~\ref{proposition: KKT-conditions} above entails that there is 
$\hat h = [\hat h _{j, \bullet}]_{j=1, \ldots, p} \in \partial \norm{\hat \theta}_{\TV, \hat w}$, 
$\hat g = [\hat g _{j, \bullet}]_{j=1, \cdots, p} \in [\partial \delta_j(\hat \theta_{j,\bullet})]_{j=1, \ldots, p}$ and 
$\hat f = [\hat f_{j,\bullet}]_{j=1, \ldots, p} \in \partial \delta_{B_d(\rho)}(\hat \theta)$ such that 
\begin{equation*}
\Big\langle \frac{1}{n} (\XB)^\top (b'(m_{\hat\theta}(\bX)) - \by) + \hat h + \hat g + \hat f, 
\hat \theta - \theta \Big\rangle = 0
\end{equation*}
for all $\theta \in \R^d$.
This can be rewritten as
\begin{align*}
&\frac 1n \inr{b'(m_{\hat\theta}(\bX))- b'(m^0(\bX)),m_{\hat \theta}(\bX) - m_\theta(\bX)}\\
&\qquad \qquad - \frac 1n \inr{\by - b'(m^0(\bX)),m_{\hat \theta}(\bX) - m_\theta(\bX)} + 
\inr{\hat h + \hat g + \hat f, \hat \theta - \theta} = 0.
\end{align*}
For any $\theta \in B_d(\rho)$ such that $n_j^\top \theta_{j, \bul} = 0$ for all $j$ and 
$h \in \partial \norm{\theta}_{ \TV, \hat w}$, the monotony of the subdifferential mapping implies 
$\inr{\hat h, \theta - \hat \theta} \leq \inr{h, \theta - \hat \theta},$  
$\inr{\hat g, \theta - \hat \theta} \leq 0,$ and $\inr{\hat f, \theta - \hat \theta} \leq 0$, so that
\begin{equation}
  \label{ineq:subdiff-monocity-logistic}
  \begin{split}
    \frac 1n \inr{b'(m_{\hat\theta}(\bX)) &- b'(m^0(\bX)),m_{\hat \theta}(\bX) - m_\theta(\bX)} \\
    &\leq \frac 1n \inr{\by - b'(m^0(\bX)),m_{\hat \theta}(\bX) - m_\theta(\bX)} - \inr{h,\hat \theta - \theta}.
  \end{split}{}
\end{equation}
Now, consider the function $H_n:\R \rightarrow \R$ defined by
\begin{equation*}
H_n(t) = \frac 1n \sum_{i=1}^n b(m_{\hat \theta + t\eta}(x_i)) - \frac{1}{n} 
\sum_{i=1}^nb'(m^0(x_i))m_{\hat \theta + t\eta}(x_i),
\end{equation*}
where $\eta$ will be defined later.
We use again the same arguments as in the proof of Lemma~\ref{lemma-connection-L2-KL}.
We differentiate $H_n$ three times with respect $t$, so that
\begin{equation*}
\begin{split}
&H'_n(t)= \frac{1}{n} \sum_{i=1}^n m_{\eta}(x_i) b'(m_{\hat \theta + t\eta}(x_i))-\frac{1}{n} \sum_{i=1}^n b'(m^0(x_i))m_{\eta}(x_i),\\
&H''_n(t) = \frac{1}{n} \sum_{i=1}^n m^2_{\eta}(x_i) b''(m_{\hat \theta + t\eta}(x_i)),\\
\text{and} \quad &H'''_n(t) = \frac{1}{n} \sum_{i=1}^n m^3_{\eta}(x_i) b'''(m_{\hat \theta + t\eta}(x_i)),
\end{split}
\end{equation*}
and in the way as in the proof of Lemma~\ref{lemma-connection-L2-KL}, we have $|H'''_n(t)| \leq C_b (C_n + \rho) 
|H''_n(t)|$, and Lemma~\ref{lemma:self-concordance} entails
\begin{equation*}
H''_n(0)\frac{ \psi(-C_b t (C_n + \rho))}{C_b^2 (C_n + \rho)^2}\leq H_n(t) - H_n(0) - tH'_n(0) \leq H''_n(0) 
\frac{ \psi(C_b t (C_n + \rho))}{C_b^2 (C_n + \rho)^2},
\end{equation*}
for all $t \geq 0$. Taking $t=1$ and  $\eta = \theta - \hat \theta$ implies 
\begin{align*}
&H_n(1)= \frac 1n \sum_{i=1}^n b(m_{\theta}(x_i))  - \frac{1}{n} \sum_{i=1}^nb'(m^0(x_i))m_{\theta}(x_i) = R(m_{\theta}),\\
\text{and} \quad & H_n(0) = \frac 1n \sum_{i=1}^n b(m_{\hat \theta}(x_i)) - \frac{1}{n} \sum_{i=1}^nb'(m^0(x_i))m_{\hat\theta}(x_i) = R(m_{\hat \theta}).  
\end{align*}
Moreover, we have
\begin{align*}
H'_n(0) &= \frac{1}{n} \sum_{i=1}^n \inr{x_i^B, \theta - \hat \theta} b'(m_{\hat \theta}(x_i))-\frac{1}{n} \sum_{i=1}^n b'(m^0(x_i))\inr{x_i^B, \hat \theta - \theta}\\
&= \frac 1n\inr{b'(m_{\hat \theta}(\bX)) - b'(m^0(\bX)),\XB(\theta - \hat \theta)},\\
\text{and} \quad H''_n(0) &= \frac{1}{n} \sum_{i=1}^n \inr{x_i^B, \hat \theta - \theta}^2 b''(m_{\hat \theta }(x_i)).
\end{align*}
Then, we deduce that
\begin{equation*}
\begin{split}
H''_n(0)\frac{ \psi(-C_b (C_n + \rho))}{C_b^2 (C_n + \rho)^2}
& \leq R(m_\theta) - R(m_{\hat \theta})  - \frac 1n\inr{b'(m_{\hat \theta}(\bX)) - b'(m^0(\bX)),\XB(\theta - \hat \theta)} \\
&=\phi \mathrm{KL}_n(m^0(\bX), m_{\theta}(\bX)) - \phi \mathrm{KL}_n(m^0(\bX), m_{\hat \theta}(\bX))\\
&\qquad + \frac 1n\inr{b'(m_{\hat \theta}(\bX)) - b'(m^0(\bX)),m_{\hat \theta}(\bX) - m_\theta(\bX)}.
\end{split}
\end{equation*}
Then, with Equation~\eqref{ineq:subdiff-monocity-logistic}, one has
\begin{equation}
\label{least-squares-in-proof}
\begin{split}
& \phi \mathrm{KL}_n(m^0(\bX), m_{\hat \theta}(\bX))+ H''_n(0)\frac{ \psi(-C_b (C_n + \rho))}{C_b^2 (C_n + \rho)^2}\\
&\qquad \leq \phi \mathrm{KL}_n(m^0(\bX), m_{\theta}(\bX))+ \frac 1n \inr{\by - b'(m^0(\bX)),m_{\hat \theta}(\bX) - m_\theta(\bX)}  - \inr{h, \hat \theta - \theta}.
\end{split}
\end{equation}
As $H''_n(0) \geq 0$, it implies that  
\begin{align}
\label{ineq:begining-oracle}
\phi \mathrm{KL}_n(m^0(\bX), m_{\hat \theta}(\bX)) &\leq \phi \mathrm{KL}_n(m^0(\bX), m_{\theta}(\bX))\nonumber\\
& \qquad+ \frac{1}{n}\inr{\by - b'(m^0(\bX)), m_{\hat \theta}(\bX) - m_\theta(\bX) } - \inr{h,\hat \theta - \theta}.
\end{align}
If $\frac{1}{n}\inr{\by - b'(m^0(\bX)), \XB(\hat\theta -  \theta)} - \inr{h, \hat \theta - \theta} < 0,$  it follows that 
\begin{equation*}
\mathrm{KL}_n(m^0(\bX), m_{\hat \theta}(\bX))\leq  \mathrm{KL}_n(m^0(\bX), m_{\theta}(\bX)),
\end{equation*}
then Theorem~\ref{thm:oracle} holds. From now on, let us assume that
\begin{equation}
\label{inequality-majoration-of-noise-2}
\frac{1}{n}\inr{\by - b'(m^0(\bX)), m_{\hat \theta}(\bX) - m_\theta(\bX)} - \inr{h, \hat \theta - \theta} \geq 0.
\end{equation}
We first derive a bound on $\frac {1}{n} \inr{\by - b'(m^0(\bX)), m_{\hat \theta}(\bX) - m_\theta(\bX)}.$ 
Recall that ${\bf{D}} ^{-1} = {\bf{T}}$ (see beginning of Section~\ref{sub:compatibility_conditions}).
We focus on finding out a bound for $\frac {1}{n} \inr{ (\XB{\bf{T}})^\top(\by - b'(m^0(\bX))), {\bf{D}}(\hat \theta - \theta)}.$
On the one hand, one has
\begin{align*}
\frac{1}{n} \inr{(\XB)^\top (\by &- b'(m^0(\bX))), \hat \theta  - \theta} \\
&= \frac{1}{n} \inr{(\XB {\bf{T}})^\top (\by - b'(m^0(\bX))), {\bf{D}}(\hat \theta - \theta)} \\
&\leq \small{\frac{1}{n} \sum_{j=1}^p \sum_{k=1}^{d_j} 
|((\XB_{\bullet,j} {T}_{j})_{\bullet, k})^\top (\by - b'(m^0(\bX))) | \; |(D_{j}({\hat \theta}_{j,\bullet} - \theta_{j,\bullet}))_k |}
\end{align*}
where $(\XB_{\bullet,j} {T}_{j})_{\bullet,k} = [ (\XB_{\bullet,j} {T}_{j})_{1,k} \cdots (\XB_{\bullet,j} {T}_{j})_{n,k}]^\top \in \R^n$ is the $k$-th column of the matrix $\XB_{\bullet,j} {T}_{j}.$ 
Let us consider the event
\begin{equation*}
\mathcal{E}_n = \bigcap_{j=1}^p \bigcap_{k =2}^{d_j}\mathcal{E}_{n,j,k}, \textrm{ where } 
\mathcal{E}_{n,j,k} = \Big\{ \frac {1}{n} | (\XB_{\bullet,j}T_j)_{\bullet,k}^\top (\by - b'(m^0(\bX))) | 
\leq \hat w_{j,k} \Big\},
\end{equation*}
so that, on $\mathcal{E}_n$, we have 
\begin{align}
\label{ineq1-proof-thm2}
\frac{1}{n} \inr{(\XB)^\top (\by - b'(m^0(\bX)), \hat \theta - \theta}  
&\leq \sum_{j=1}^p \sum_{k=1}^{d_j} \hat w_{j,k} |(D_{j}({\hat \theta}_{j,\bullet} - \theta_{j,\bullet}))_k| \nonumber \\
& \leq   \sum_{j=1}^p  \norm{ \hat w_{j,\bullet} \odot D_{j}({\hat \theta}_{j,\bullet} - \theta_{j,\bullet})}_1.
\end{align}
On the other hand, from the definition of the subgradient $[h_{j, \bullet}]_{j=1, \ldots, p} \in \partial \norm{\theta}_{\TV, \hat w}$ (see Equation~\eqref{subdifferential-of-TVw}), one can choose $h$ such that 
\begin{equation*}
h_{j, k} = (D_{j}^\top (\hat w_{j,\bullet}\odot\sgn(D_{{j}}{\theta}_{j,\bullet})))_k
\end{equation*}
for all $k \in J_j(\theta)$ and
\begin{equation*}
h_{j, k} = (D_{j}^\top (\hat w_{j,\bullet}\odot\sgn ( D_{j}\hat \theta_{j, \bullet} ) )_k = (D_{j}^\top (\hat w_{j,\bullet}\odot\sgn ( D_{j}(\hat \theta_{j, \bullet} - \theta_{j,\bullet}) ) )_k
\end{equation*}
for all $k \in J_j^\complement(\theta)$.
Using a triangle inequality and the fact that $\sgn(x)^\top x= \norm{x}_1$, we obtain
\begin{align}
\label{ineq2-proof-thm1}
-\inr{h, \hat \theta - \theta} 
&\leq \sum_{j=1}^p \norm{ (\hat w_{j,\bullet})_{J_j(\theta)} \odot D_{{j}}(\hat \theta_{j, \bullet} -\theta_{j, \bullet})_{J_j(\theta)}}_1 \nonumber\\
& \quad - \sum_{j=1}^p \norm{ (\hat w_{j,\bullet})_{J^\complement_j(\theta)} \odot D_{{j}}(\hat \theta_{j, \bullet} -\theta_{j, \bullet})_{J^\complement_j(\theta)}}_1 \nonumber\\
&\leq  \sum_{j=1}^p \norm{(\hat \theta_{j, \bullet} -\theta_{j, \bullet})_{J_j(\theta)}}_{\TV, \hat w_{j,\bullet}} - \sum_{j=1}^p \norm{ (\hat \theta_{j, \bullet} -\theta_{j, \bullet})_{J^\complement_j(\theta)}}_{\TV, \hat w_{j,\bullet}}.
\end{align}
Combining inequalities~\eqref{ineq1-proof-thm2} and~\eqref{ineq2-proof-thm1}, we get
\begin{equation*}
\sum_{j=1}^p \norm{ (\hat \theta_{j, \bullet} -\theta_{j, \bullet})_{J^\complement_j(\theta)}}_{ \TV, \hat w_{j,\bullet}} \leq 2\sum_{j=1}^p  \norm{(\hat \theta_{j, \bullet} -\theta_{j, \bullet})_{ J_j(\theta)}}_{ \TV, \hat w_{j,\bullet}}
\end{equation*}
on $\mathcal{E}_n$.
Hence
\begin{equation*}
\sum_{j=1}^p \norm{(\hat w_{j,\bullet})_{J^\complement_j(\theta)} \odot D_{{j}}(\hat \theta_{j, \bullet} -\theta_{j, \bullet})_{J^\complement_j(\theta)}}_1 \leq 2\sum_{j=1}^p \norm{(\hat w_{j,\bullet})_{ J_j(\theta)} \odot D_{{j}}(\hat \theta_{j, \bullet} -\theta_{j, \bullet})_{J_j(\theta)}}_1.
\end{equation*}
This means that 
\begin{equation}
\label{Delta-hat-theta-in-Cone-square}
\hat \theta - \theta \in \mathscr{C}_{\TV, \hat w}(J(\theta)) \textrm{  and } {\bf{D}}(\hat \theta - \theta) \in \mathscr{C}_{1,\hat w}(J(\theta)),
\end{equation}
see~\eqref{C-AGG} and~\eqref{C-1}.
Now, going back to~\eqref{ineq:begining-oracle} and taking into account~\eqref{Delta-hat-theta-in-Cone-square}, the compatibility of $\XB {\bf{T}}$ given in Equation~\eqref{ineq:compatibility-XT} provides the following on the event  $\mathcal{E}_n$:
\begin{equation*}
\begin{split}
\phi \mathrm{KL}_n(m^0(\bX), m_{\hat \theta}(\bX))
&\leq \phi \mathrm{KL}_n(m^0(\bX), m_\theta(\bX)) \\
&\quad + 2\sum_{j=1}^p  \norm{ (\hat w_{j,\bullet})_{J_j(\theta)} \odot D_{j}(\hat \theta_{j, \bullet} -\theta_{j, \bullet})_{J_j(\theta)}}_1.
\end{split}
\end{equation*}
Then
\begin{equation}
\label{for-case-least-squares}
\begin{split}
\mathrm{KL}_n(m^0(\bX), m_{\hat \theta}(\bX))&\leq \mathrm{KL}_n(m^0(\bX), m_\theta(\bX)) + \frac{\norm{m_{\hat \theta}(\bX) - m_\theta(\bX) }_2}{ \sqrt{n}\, \phi\, \kappa_{\bf{T},\hat \gamma}(J(\theta)) \kappa(J(\theta))},
\end{split}
\end{equation}
where $\hat \gamma = (\hat \gamma_{1,\bullet}^\top, \ldots, \hat \gamma_{p,\bullet}^\top)^\top$ is such that 
\begin{equation*}
\hat \gamma_{j,k} = 
 \left\{
   \begin{array}{ll}
    2\hat {w}_{j,k} & \mbox{if }  k \in J_j(\theta),\\
     0  & \mbox{if }   k \in J_j^\complement(\theta),
    \end{array} 
\right.
\end{equation*}
for all $j=1, \ldots, p$ and 
\begin{equation*}
\kappa_{{\bf{T}}, \hat \gamma}(  J(\theta)) =\bigg\{ 32  \sum_{j=1}^p\sum_{k=1}^{d_j} |\hat \gamma_{j,k+1} - \hat \gamma_{j,k}|^2 + 2|J_j(\theta)| \norm{\hat \gamma_{j,\bullet}}_\infty^2\Delta_{\min, J_j(\theta)}^{-1}\bigg\}^{-1/2}.
\end{equation*}
Now, we find an upper bound for
\begin{equation*}
\frac{1}{ \kappa^2_{{\bf{T}},\hat\gamma}(J(\theta))}
= 32\sum_{j=1}^p\sum_{k=1}^{d_j} |\hat \gamma_{j,k+1} - \hat \gamma_{j,k}|^2 + 2|J_j(\theta)| \norm{\hat \gamma_{j,\bullet}}_\infty^2\Delta_{\min, J_j(\theta)}^{-1}.
\end{equation*}
Note that $\norm{\hat \gamma_{j,\bullet}}_\infty \leq 2\norm{\hat w_{j,\bullet}}_\infty$. 
Let us write $J_j(\theta) =\big\{k_j^1, \ldots, k_j^{|J_j(\theta)|}\big\}$ and set $B_r =[\![k_j^{r-1}, k_j^{r}[\![ \; = \{k_j^{r-1}, k_j^{r-1} + 1, \ldots, k_j^{r} -1\}$ for $r = 1, \ldots, |J_j(\theta)|+1$ with the convention that $k_j^0=0$ and $k_j^{|J_j(\theta)|+1} = d_j+1$. Then
\begin{equation*}
\begin{split}
\sum_{k=1}^{d_j} |\hat \gamma_{j,k+1} -\hat \gamma_{j,k}|^2 &= \sum_{r=1}^{|J_j(\theta)|+1} \sum_{k \in B_r} |\hat\gamma_{j,k+1} -\hat\gamma_{j,k}|^2\\
& =\sum_{r=1}^{|J_j(\theta)|+1} |\hat \gamma_{j,k_j^{r -1}+1} - \hat\gamma_{j,k_j^{r -1}}|^2 + |\hat \gamma_{j,k_j^{r}} - \hat \gamma_{j,k_j^{r }-1}|^2 \\
& =\sum_{r=1}^{|J_j(\theta)|+1} \hat \gamma_{j,k_j^{r -1}}^2 + \hat \gamma_{j,k_j^{r}}^2 \\
& = \sum_{r=1}^{|J_j(\theta)|} 2\ \hat \gamma_{j,k_j^{r}}^2\\
& \leq 8\ |J_j(\theta)|\ \norm{(\hat w_{j,\bullet})_{J_j(\theta)}}_\infty^2.
\end{split}
\end{equation*}
Therefore
\begin{equation}
\label{eq:upper_bound_kappa_2_T}
\begin{split}
\frac{1}{\kappa^2_{{\bf{T}},\hat \gamma}(J(\theta))}
&\leq 512 \sum_{j=1}^p \Big( |J_j(\theta)|\ \norm{(\hat w_{j,\bullet})_{J_j(\theta)}}_\infty^2 +  |J_j(\theta)|\ \norm{(\hat w_{j,\bullet})_{J_j(\theta)}}_\infty^2\Delta_{\min, J_j(\theta)}^{-1} \Big) \\
& \leq 512 \sum_{j=1}^p \Big( 1 + \frac{1}{\Delta_{\min, J_j(\theta)}} \Big) |J_j(\theta)| \norm{(\hat w_{j,\bullet})_{J_j(\theta)}}_\infty^2\\
& \leq 512 |J(\theta)| \max_{j=1, \ldots, p}\norm{(\hat w_{j,\bullet})_{J_j(\theta)}}_\infty^2.
\end{split}
\end{equation}
Now, we use the connection between the empirical norm and Kullback-Leibler divergence.
Indeed, using Lemma~\ref{lemma-connection-L2-KL}, we get
\begin{align*}
&\frac{\norm{ m_{\hat \theta}(\bX) - m_\theta(\bX) }_2}{ \sqrt{n}\phi \kappa_{{\bf{T}},\hat\gamma}(  J(\theta)) \kappa(J(\theta))} \\
&\quad  \quad \leq
 \frac{1}{\sqrt{\phi} \kappa_{{\bf{T}},\hat\gamma}(J(\theta)) \kappa(J(\theta))}\Big(\frac{1}{\sqrt{n}} 
 \norm{ m_{\hat \theta}(\bX)- m^0(\bX)}_2 + \frac{1}{\sqrt{n}}\norm{ m^0(\bX) - m_{\theta}(\bX)}_2\Big) \\
&\quad \quad \leq \frac{2}{\sqrt{\phi}\kappa_{{\bf{T}},\hat\gamma}(J(\theta)) 
\kappa(J(\theta)) \sqrt{C_n(\rho, L_n)} } \Big(\mathrm{KL}_n(m^0(\bX), m_{\hat \theta}(\bX))^{1/2} \\
&\qquad \qquad \qquad \qquad \qquad \qquad \qquad \qquad \qquad \qquad 
+ \mathrm{KL}_n(m^0(\bX), m_{\theta}(\bX))^{1/2} \Big),
\end{align*}
where we defined $C_n(\rho, L_n) = \frac{L_n \psi(-C_b (C_n + \rho))}{C_b^2 \phi (C_n + \rho)^2}$, so that combined with
Equation~\eqref{for-case-least-squares}, we obtain
\begin{align*}
  \mathrm{KL}_n(m^0(\bX), m_{\hat \theta}(\bX)) &\leq \mathrm{KL}_n(m^0(\bX), m_\theta(\bX)) \\
  & \quad + \frac{2}{\sqrt{\phi}\kappa_{{\bf{T}},\hat\gamma}(J(\theta)) \kappa(J(\theta)) \sqrt{C_n(\rho, L_n)}} 
  \big(\mathrm{KL}_n(m^0(\bX), m_{\hat \theta}(\bX))^{1/2} \\
  & \qquad \qquad \qquad \qquad \qquad \qquad \qquad \qquad \quad 
  + \mathrm{KL}_n(m^0(\bX), m_{\theta}(\bX))^{1/2} \big).
\end{align*}
This inequality entails the following upper bound
\begin{equation*}
  \mathrm{KL}_n(m^0(\bX), m_{\hat \theta}(\bX)) \leq 3 \mathrm{KL}_n(m^0(\bX), m_\theta(\bX)) 
  + \frac{5}{\phi \kappa^2_{{\bf{T}},\hat\gamma}(J(\theta)) \kappa^2(J(\theta)) C_n(\rho, L_n)},
\end{equation*}
since whenever we have $x \leq c + b \sqrt x$ for some $x, b, c > 0$, then $x \leq 2c + b^2$.
Introducing $g(x) = x^2 / \psi(-x) = x^2 / (e^{-x} + 1 - x)$, we note that 
\begin{equation*}
  \frac{1}{C_n(\rho, L_n)} = \frac{\phi}{L_n} g(C_b (C_n + \rho)) \leq \frac{\phi}{L_n} 
  (C_b(C_n + \rho) + 2),
\end{equation*}
since $g(x) \leq x + 2$ for any $x > 0$.
Finally, by using also~\eqref{eq:upper_bound_kappa_2_T}, we end up with
\begin{equation*}
  \mathrm{KL}_n(m^0(\bX), m_{\hat \theta}(\bX)) \leq 3 \mathrm{KL}_n(m^0(\bX), m_\theta(\bX)) 
  + \frac{2560 (C_b(C_n + \rho) + 2)}{L_n \kappa^2(J(\theta))} \; |J(\theta)| \; 
  \norm{(\hat w_{j,\bullet})_{J_j(\theta)}}_\infty^2,
\end{equation*}
which is the statement provided in Theorem~\ref{thm:oracle}.
The only thing remaining is to control the probability of the event $\mathcal{E}_n^\complement$. 
This is given by the following:
 \begin{equation*}
\begin{split}
\P[\mathcal{E}_n^\complement] &\leq \sum_{j=1}^p \sum_{k=2}^{d_j} 
\P \Big[\frac{1}{n} | (\XB_{\bullet,j} T_{j})_{\bullet,k}^\top (\by - b'(m^0(\bX)))| \geq \hat w_{j,k} \Big] \\
&\leq \sum_{j=1}^p \sum_{k=2}^{d_j} \P \Big[\sum_{i=1}^n  |(\XB_{\bullet,j} T_{j})_{i,k}(y_i - b'(m^0(x_i))) | \geq { n\hat w_{j,k}} \Big].
\end{split}
\end{equation*}
Let $\xi_{i,j,k} = (\XB_{\bullet,j} {T}_{j})_{i,k}$ and $Z_{i}= y_i - b'(m^0(x_i)).$
Note that conditionally on $x_i$, the random variables $(Z_{i})$
are independent.
It can be easily shown (see Theorem 5.10 in~\cite{lehmann1998}) that the moment generating function of $Z$ (copy of $Z_i$) is given by
\begin{equation}
\label{moment-gener-func}
\E[\exp(tZ)] = \exp(\phi^{-1}\{b(m^0(x) + t) - tb'(m^0(x) - b(m^0(x)))\}).
\end{equation}
Applying Lemma 6.1 in~\cite{rigollet2012}, using~\eqref{moment-gener-func} and Assumption~\ref{ass:glm}, we can derive the following Chernoff-type bounds 
\begin{equation}
\label{cal-proba-Ecomp}
\P\Big[ \sum_{i=1}^n |\xi_{i,j,k}Z_{i}| \geq n\hat w_{j,k}\Big]
 \leq 2\exp\Big(- \frac{n^2\hat w^2_{j,k} }{2U_n\phi\norm{\xi_{\bullet,j,k}}_2^2}\Big),
\end{equation}
where $\xi_{\bullet,j,k} = [\xi_{1,j,k} \cdots \xi_{n,j,k} ]^\top \in \R^n.$
We have 
\begin{equation*}
\bX^B_{\bullet,j} {T}_{j} =
\begin{bmatrix}
1 & \sum_{k=2}^{d_j} x_{1,j,k}^B & \sum_{k=3}^{d_j} x_{1,j,k}^B  & \cdots&  
\sum_{k=d_{j-1}}^{d_j} x_{1,j,k}^B &  x_{1,j,d_j}^B \\
\vdots & \vdots & \vdots&  & \vdots &\vdots \\
1 &\sum_{k=2}^{d_j} x_{n,j,k}^B & \sum_{k=3}^{d_j} x_{n,j,k}^B  & \cdots&  
 \sum_{k=d_{j-1}}^{d_j} x_{n,j,k}^B & x_{n,j,d_j}^B
\end{bmatrix},
\end{equation*}
therefore
\begin{equation}
\label{eq:empirical-norm-XT}
\norm{\xi_{\bullet,j,k}}_2^2 = \sum_{i=1}^n(\bX^B_{\bullet,j} {T}_{j})^2_{\bullet, k} =  \# \Big(\Big\{ i : x_{i,j} \in \bigcup_{r=k}^{d_j} I_{j,r} \Big\} \Big) = n\hat \pi_{j,k}.
\end{equation}
So, using the weights $\hat w_{j,k}$ given by~\eqref{choice-of-weights-sq-slow-GLM} together with~\eqref{cal-proba-Ecomp} and~\eqref{eq:empirical-norm-XT}, we obtain that the probability of $\mathcal{E}_{n}^\complement$ is smaller than $2e^{-A}.$ 
This concludes the proof of the first part of Theorem~\ref{thm:oracle}.
 $\hfill \square$

\subsection{Proof of Theorem~\ref{thm:additive}} 
\label{sub:proof_of_theorem_thm:additive}

First, let us note that in the least squares setting, we have $R(m_{\theta}) - R(m^0) = \norm{m_{\theta} - m^0}_n^2$ for any $\theta \in \R^d$ where $\norm{g}_n^2 = 
\frac 1n \sum_{i=1}^n g(x_i)^2$, and that $b(y) = \frac 12 y^2$, $\phi = \sigma^2$ (noise variance) in Equation~\eqref{distribut-glm}, and $L_n = U_n = 1$, $C_b = 0$.
Theorem~\ref{thm:oracle} provides 
\begin{equation*}
  \norm{m_{\hat \theta} - m^0}_n^2 \leq 3 \norm{m_{\theta} - m_{\theta^0}}_n^2 
  + \frac{5120 \sigma^2}{\kappa^2(J(\theta))} \frac{|J(\theta)|(A + \log d)}{n}
\end{equation*}
for any $\theta \in \R^d$ such that $n_j^\top \theta_{j \bul} = 0$ and $J(\theta) \leq J_*$.
Since $d_j = D$ for all $j=1, \ldots, p$, we have $d = D p$ and
\begin{equation}
  \label{eq:sparsity-comparison}
  | J(\theta) | = \sum_{j=1}^p | \{ k = 2, \ldots, D : \theta_{j, k} \neq \theta_{j, k-1 } \} | \leq (D - 1) |\cJ(\theta)| \norm{\theta}_\infty \leq D p \norm{\theta}_\infty
\end{equation}
for any $\theta \in \R^d$, where we recall that $\cJ(\theta) = \{ j=1, \ldots, p : \theta_{j, \bul} \neq \boldsymbol 0_{D} \}$.
Also, recall that $I_{j, 1} = I_1 = [0, \frac{1}{D}]$ and $I_{j, k} = I_k = (\frac{k-1}{D}, \frac{k}{D}]$ for $k=2, \ldots, D$
and $j = 1, \ldots, p$.
Also, we consider $\theta = \theta^*$, where $\theta_{j, \bul}^*$ is defined, for any $j \in \cJ_*$, as the minimizer of
\begin{equation*}
  \sum_{i=1}^n \Big( \sum_{k=1}^{D} (\theta_{j, k} - m_j^0(x_{i, j}) ) 
  \ind{I_{k}}(x_{i, j}) \Big)^2
\end{equation*}
over the set of vectors $\theta_{j, \bul} \in \R^{D}$ satisfying $n_j^\top \theta_{j, \bul} =  0$, and we put
$\theta_{j, \bul}^* = \boldsymbol 0_D$ for $j \notin \cJ_*$. 
It is easy to see that the solution is given by
\begin{equation*}
  \theta_{j, k}^* = \frac{\sum_{i=1}^n m_j^0(x_{i, j}) \ind{I_{k}}(x_{i, j})}{n_{j, k}},
\end{equation*}
where we recall that $n_{j, k} = \sum_{i=1}^n \ind{I_{k}}(x_{i, j})$.
Note in particular that the identifiability assumption $\sum_{i=1}^n m_j^0(x_{i, j}) = 0$ entails that $n_j^\top \theta_{j, \bul}^* = 0$.
In order to control the bias term, an easy computation gives that, whenever $x_{i, j} \in I_{k}$
\begin{equation*}
  |\theta_{j, k}^*  - m_j^0(x_{i, j})| \leq 
  \frac{\sum_{i'=1}^n | m_j^0(x_{i', j}) - m_j^0(x_{i, j}) | 
  \ind{I_{k}}(x_{i', j})}{n_{j, k}} \leq L |I_{k}| = \frac{L}{D},
\end{equation*}
where we used the fact that $m_j^0$ is $L$-Lipschitz, so that
\begin{align*}
  \norm{m_{\theta^*} - m^0}_n^2 &= \frac 1n \sum_{i=1}^n 
  (m_{\theta^*}(x_{i, j}) - m^0(x_{i, j}))^2 \\
  &= \frac 1n \sum_{i=1}^n \Big( \sum_{j \in \cJ_*} 
  \sum_{k=1}^{D} (\theta_{j, k}^* - m^0(x_{i, j})) \ind{I_{k}} \Big)^2 \\
  &\leq \frac{|\cJ_*|}{n} \sum_{i=1}^n \Big( \sum_{j \in \cJ_*} 
  \sum_{k=1}^{D} (\theta_{j, k}^* - m^0(x_{i, j})) \ind{I_{k}} \Big)^2 \\
  &\leq \frac{|\cJ_*|}{n} \sum_{i=1}^n \sum_{j \in \cJ_*} 
  \sum_{k=1}^{D} (\theta_{j, k}^* - m^0(x_{i, j}))^2 \ind{I_{k}}(x_{i, j}) \\
  &\leq |\cJ_*| \sum_{j \in \cJ_*} 
  \sum_{k=1}^{D} L^2 |I_{k}|^2 \ind{I_{k}}(x_{i, j})
  \leq \frac{L^2 |\cJ_*|^2}{D^2}.
\end{align*}
Note that $|\theta_{j, k}^*| \leq \norm{m_j^0}_{n, \infty}$ where $\norm{m_j^0}_{n, \infty} = \max_{i=1, \ldots, n} |m_j^0(x_{i, j})|$. This entails that
 $\norm{\theta^*}_\infty \leq \max_{j=1, \ldots, p} \norm{m_j^0}_{n, \infty} = M_n$.
So, using also~\eqref{eq:sparsity-comparison}, we end up with
\begin{align*}
  \norm{m_{\hat \theta} - m^0}_n^2 &\leq \frac{3 L^2 |\cJ_*|^2}{D^2} 
  + \frac{5120 \sigma^2}{\kappa^2(J(\theta^*))} \frac{D \cJ_* M_n (A 
  + \log (D p M_n))}{n},
\end{align*}
which concludes the proof Theorem~\ref{thm:additive} using $D = n^{1/3}$.  $\hfill \square$


\end{document}